\documentclass[letterpaper]{article} %
\usepackage{aaai25}  %
\usepackage{times}  %
\usepackage{helvet}  %
\usepackage{courier}  %
\usepackage[hyphens]{url}  %
\usepackage{graphicx} %
\usepackage{natbib}  %
\usepackage{caption} %
\usepackage{algorithm}
\usepackage{algorithmic}

\usepackage{newfloat}
\usepackage{listings}

\usepackage{subcaption}
\usepackage{amssymb}
\usepackage{microtype}
\usepackage{url}
\usepackage{booktabs}
\usepackage{graphicx} 
\usepackage{amsmath}
\usepackage{multirow}
\usepackage{makecell}
\usepackage{listings}
\usepackage{color}
\usepackage{xcolor}
\usepackage{tabularx}  

\DeclareCaptionStyle{ruled}{labelfont=normalfont,labelsep=colon,strut=off} %
\floatstyle{ruled}
\newfloat{listing}{tb}{lst}{}
\floatname{listing}{Listing}

\newcommand\our{\textsc{Kosmos-2.5}}
\newcommand\ourchat{\textsc{Kosmos-2.5-chat}}
\title{\our{}: A Multimodal Literate Model}

\author{\vspace{-0.25in} \\
\bf Tengchao Lv\thanks{~Equal contribution. $\dagger$ Corresponding author.},~~Yupan Huang\footnotemark[1],~~Jingye Chen\footnotemark[1],~~Yuzhong Zhao, ~~Yilin Jia, ~~Lei Cui$^\dagger$,\\~~Shuming Ma,~~Yaoyao Chang,~~Shaohan Huang,~~Wenhui Wang,~~Li Dong,\\~~Weiyao Luo,~~Shaoxiang Wu,~~Guoxin Wang,~~Cha Zhang,~~Furu Wei\footnotemark[2] \\
\vspace{1mm}
Microsoft \\\
{\href{https://aka.ms/GeneralAI}{aka.ms/GeneralAI}}
\vspace{-0.7cm}
\\}

\usepackage{hyperref}  
\usepackage{bibentry}

\begin{document}

\maketitle

\begin{abstract}
The automatic reading of text-intensive images represents a significant advancement toward achieving Artificial General Intelligence (AGI). In this paper we present \our{}, a multimodal literate model for machine reading of text-intensive images. Pre-trained on a large-scale corpus of text-intensive images, \our{} excels in two distinct yet complementary transcription tasks: (1) generating spatially-aware text blocks, where each block of text is assigned spatial coordinates within the image, and (2) producing structured text output that captures both style and structure in markdown format. This unified multimodal literate capability is achieved through a shared decoder-only autoregressive Transformer architecture and task-specific prompts. 
Building on this foundation, we fine-tune \our{} for document understanding tasks, resulting in a document understanding generalist named \ourchat{}. 
Additionally, a large corpus of 357.4 million document pages spanning diverse domains was curated for pre-training.
We evaluate \our{} on two newly proposed benchmarks, OCREval and MarkdownEval, for document-level text recognition and image-to-markdown generation, demonstrating impressive literate capabilities comparable to GPT-4o.
\ourchat{} achieves performance comparable to other state-of-the-art generalists that are five times larger (1.3B vs. 7B) across nine text-rich visual question answering benchmarks.
Models and code have been available at \url{https://aka.ms/kosmos25}.
\end{abstract}

\begin{figure*}[t]
\centering
\small
\includegraphics[width=1.\textwidth]{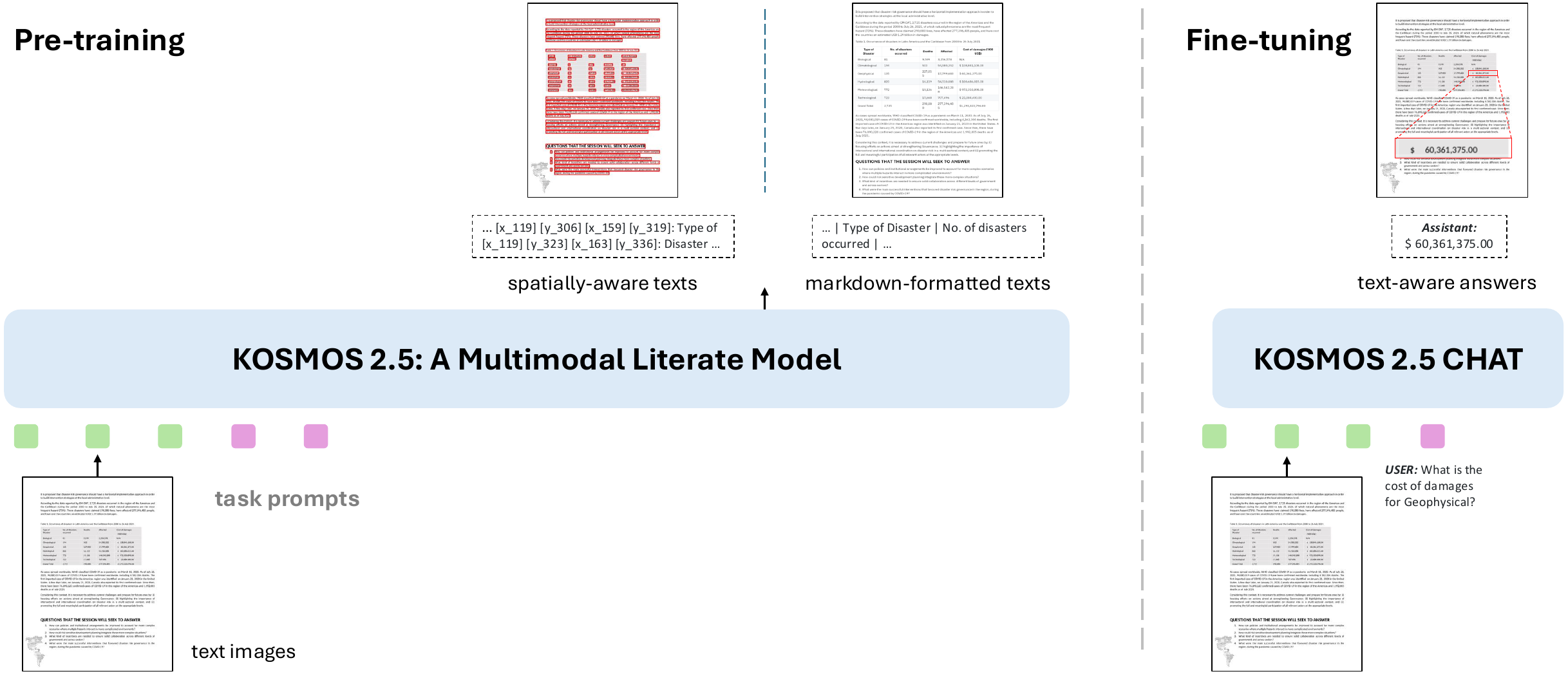}
\caption{
\our{} is a multimodal document foundation model that takes text images as input and generates spatially-aware texts (i.e., texts with bounding boxes) or markdown-formatted texts (i.e., texts with markdown elements), following different task prompts, respectively. The model possesses the ability to comprehensively perceive textual content, its spatial context, and nuances of formatting and style within a unified framework. \ourchat{} is fine-tuned from \our{}. It is a visual document understanding generalist that can answer user-provided questions about text-rich images from various domains.
}
\label{fig:introduction}
\end{figure*}

\section{Introduction}
Multimodal large language models (MLLMs) extend the capabilities of large language models (LLMs) to multimodal tasks, enabling them to process and generate responses from both textual and visual inputs~\cite{zhang2023llavar,liu2024visual,chatgpt,touvron2023llama}. However, while existing MLLMs have primarily focused on natural images, the challenge of effectively reading and understanding text-intensive images—such as academic papers, receipts, design documents, and web pages—remains underexplored.

Traditional Optical Character Recognition (OCR) methods are primarily designed for generating line-level text content and capturing its spatial positions within an image. Although these methods preserve layout information, they often neglect the document-level reading order and structural integrity that are crucial for accurate document understanding. On the other hand, markdown-formatted text offers significant advantages over plain text by explicitly distinguishing between different structural elements—such as tables, lists, and headings—through specific tokens. Current approaches are either limited to line-level text recognition~\cite{mplug-docowl,mplug-docowl-1.5,li2023monkey} or focus on structured parsing within a specific document category~\cite{blecher2023nougat}, making it difficult to achieve comprehensive document-level reading and understanding capabilities across diverse categories.

Motivated by these observations, we present \textbf{\our{}}, a multimodal literate model designed to address the unique challenges of reading and understanding text-intensive documents, including capturing the reading order and structural integrity of the content.
As illustrated in Figure~\ref{fig:introduction}, \our{} is pre-trained on two distinct yet complementary generative tasks: document-level text recognition and image-to-markdown generation. The first task involves generating spatially-aware text blocks, assigning text lines to their corresponding spatial coordinates within the original text-rich image. The second task focuses on producing structured text output that captures both style and structure in markdown format. Both tasks are performed within a unified framework using task-specific prompts, leveraging a shared Transformer architecture that combines a ViT-based vision encoder and a Transformer-based language decoder connected by a resampler module~\cite{vit,lee2023pix2struct,alayrac2022flamingo}.

To realize the potential of our pre-trained model and validate its effectiveness in downstream understanding tasks, we further fine-tune \our{} for document understanding tasks, resulting in \textbf{\ourchat{}}, which can answer user-provided questions about text-rich images. Despite having only 1.3B parameters, \ourchat{} achieves performance comparable to other state-of-the-art generalists with over 7B parameters on various text-rich visual question answering benchmarks.

Given the absence of a comprehensive document reading dataset, we curated a large corpus of \textbf{357.4 million document pages}, including scanned documents, general documents, academic papers, web pages, design images, handwritten texts, mathematical content, and project documents. Each document is annotated with text lines with bounding boxes or markdown formats. This dataset was constructed using an automatic pipeline for data collection, filtering, and quality control, offering valuable insights for future research.

Existing document reading benchmarks primarily focus on line-level text reading capabilities~\cite{Liu2023ocrbench} or are limited to specific domains, such as converting academic papers to markdown format \cite{blecher2023nougat}. To comprehensively evaluate models’ capabilities in document-level text recognition and image-to-markdown generation tasks, we introduce two extensive benchmarks: \textbf{OCREval} and \textbf{MarkdownEval}. Specifically, OCREval contains 2,297 samples, while MarkdownEval includes 5,633 samples. The benchmarks cover a diverse range of document categories, including handwritten texts, design documents, receipts, academic papers, web pages, mathematical content, tables, and more. Experimental results on these benchmarks demonstrate that \our{} exhibits impressive literate capabilities on par with GPT-4o~\cite{gpt4}.

The contributions of this work are summarized as follows:

\begin{itemize}

\item {We propose two distinct yet cooperative document reading tasks for pre-training a foundational document model capable of machine reading and understanding the order and structure of text-intensive documents. The pre-trained \our{} demonstrates impressive multimodal literate capabilities on par with GPT-4o, and the fine-tuned \ourchat{} achieves competitive results across nine document understanding benchmarks.}

\item {We curated a large and diverse corpus consisting of 357.4 million text-rich document images, with text lines annotated with bounding boxes or in markdown format. The automated data curation pipeline provides valuable insights for future research.}

\item {We introduce two comprehensive benchmarks, OCREval and MarkdownEval, to provide thorough evaluations of document-level machine reading capabilities.}

\end{itemize}

\begin{figure*}[ht]
\centering
\includegraphics[width=1.0\textwidth]{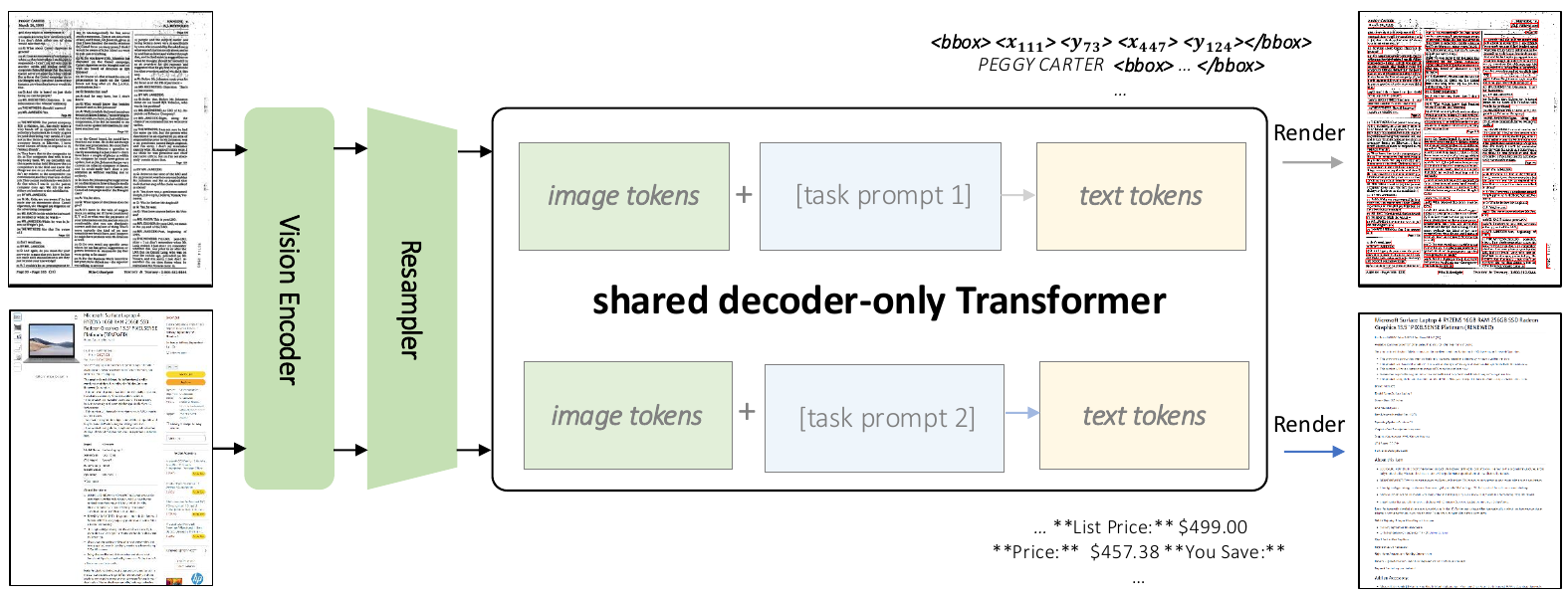}
\caption{
The model architecture of \our{} leverages a shared Transformer architecture that combines a ViT-based vision encoder and a Transformer-based language decoder connected by a resampler module.
}
\label{fig:model_arch}
\end{figure*}

\begin{table*}[t]
\scriptsize
\centering
\setlength{\tabcolsep}{2mm}
\begin{tabular}{cccl}
\toprule
\textbf{Training Stage} & \textbf{Task} & \textbf{Definition} & \textbf{Prompt}\tabularnewline
\midrule
\multirow{4}{*}{\textbf{Pre-training}} & \multirow{2}{*}{\textbf{\makecell{Document-level  \\Text Recognition}}} & \multirow{2}{*}{\makecell{Generating spatially-aware text blocks, where each block\\ of text is assigned its spatial coordinates within the image}} & \texttt{<s><image>Image Embedding</image><ocr>}\tabularnewline
& & & \textcolor[rgb]{0,0.4,0}{$\bigcup_{n=1}^{N}$ ($\mathbf{B}_n \oplus \mathbf{T}_n)$ \texttt{</s>}}\tabularnewline
\cline{2-4}
& \multirow{2}{*}{\textbf{\makecell{Image-to-Markdown \\Generation}}} & \multirow{2}{*}{\makecell{Producing structured text output that captures \\ styles and structures into the markdown format}} & \texttt{<s><image>Image Embedding</image><md>}\tabularnewline
& & & \textcolor[rgb]{0,0.4,0}{[Markdown Text]\texttt{</s>}}\tabularnewline
\midrule
\multirow{5}{*}{\textbf{Fine-tuning}} & \multirow{5}{*}{\textbf{\makecell{Document \\Understanding}}} & \multirow{4}{*}{\makecell{\\Answering the user-provided text-related questions \\ about text-intensive images}} & \texttt{<s><image>Image Embedding</image><md>}\tabularnewline
& & & A chat between a curious user and an artificial intelligence\tabularnewline
& & & assistant. The assistant gives helpful, detailed, and polite\tabularnewline
& & & answers to the user's questions. USER: [Question]\tabularnewline
& & & ASSISTANT: \textcolor[rgb]{0,0.4,0}{[Answer]\texttt{</s>}}\tabularnewline
\bottomrule
\end{tabular}

\caption{Tasks, prompts, and \textcolor[rgb]{0,0.4,0}{response sequence} formats used to train \our{}. Special tokens \texttt{<s>} and \texttt{</s>} denote sequence boundaries, while \texttt{<image>} and \texttt{</image>} indicate the start and end of image embeddings. For document-level text recognition tasks, the operator \(\oplus\) represents the concatenation of the text line \(\mathbf{T}_n\) and its bounding box \(\mathbf{B}_n\). During pre-training, special tokens \texttt{<ocr>} and \texttt{<md>} denote document-level text recognition and text-to-markdown generation tasks, respectively. For visual document understanding tasks, we use the same format as text-to-markdown generation tasks since these do not require bounding box outputs.}
\label{tab:task_prompts}
\end{table*}

\begin{table*}[t]
\scriptsize
\centering

\begin{tabular}{lllcc}
\toprule
\bf Format and Task & \bf Document Category & \bf Description & \bf \makecell{Page \\Num} & \bf \makecell{Sampling\\Ratio} \\
\midrule
\multirow{6}{*}{\makecell{\\ \\ \\Layout-based \\(texts+bboxes) \\Document-level \\Text Recognition}}
&Scanned Document & 
\makecell[l]{Includes IIT-CDIP~\cite{lewis2006building}, a large collection of scanned documents.}
& 27.6M & 10\% \\
\cline {2-5}  
&General Document & 
\makecell[l]{Includes general PDFs and SEC files. General PDFs are crawled from the web, resulting in a\\diverse open-domain digital PDF corpus. SEC files are sourced from SEC.gov and comprise\\various companies' periodic reports, filings, and forms.}
& 187.4M & 20\% \\
\cline {2-5}  
&Academic Paper &
Includes arXiv papers.
& 20.9M & 5\% \\
\cline {2-5}  
&Web Page &
Self-constructed large-scale dataset of crawled web pages.
& 100.5M & 10\% \\
\cline {2-5}  
&Design Image & 
\makecell[l]{Includes PowerPoint, posters, and MARIO-10M\cite{chen2024textdiffuser} collected from various sources.}
& 6.2M & 3\% \\
\cline {2-5}  
&Handwritten Image&
\makecell[l]{Includes Synt-handwritten data produced using a wide range of handwritten font files.}
& 0.2M & 1\% \\
\cline {2-5} 
& Math Image &
\makecell[l]{Includes CROHME\cite{mouchere2014icfhr} and IM2LATEX-100K\cite{deng2017image},\\CROHME contains various handwritten mathematical expressions. IM2LATEX-100K is a large\\dataset containing mathematical expressions with corresponding LaTeX markup.
}
& 0.6M & 1\% \\
\midrule
\multirow{4}{*}{\makecell{\\Markup-based \\(texts+markdown) \\Image-to-Markdown \\Generation}}
&General Document &
\makecell[l]{Includes Docx type files and SEC files sourced from SEC.gov. They are crawled from the web \\and converted tomarkdown format. Each page corresponds to its markdown information. }
& 1.1M & 10\% \\
\cline {2-5}  
&Academic Paper &
\makecell[l]{A subset of the entire arXiv papers is used to extract the mapping of PDF pages and\\its corresponding markdown information converted from the LATEX code.}
& 3.7M & 15\% \\
\cline {2-5}  
&Project Document &
\makecell[l]{Includes “README.md” files of open-source GitHub projects, primarily in markdown format.}
& 2.9M & 15\% \\
\cline {2-5}  
&Web Page &
\makecell[l]{Self-constructed large-scale dataset of crawled web pages, and its corresponding\\markdown information converted from the HTML code.}
& 6.3M & 10\% \\
\midrule
\multicolumn{3}{c}{\textbf{Total}}  & 357.4M & 100\% \\
\bottomrule
\end{tabular}

\caption{Summary of data used to pre-train \our{}, including descriptions of each document category, the number of pages, and their respective sampling ratios in the training data.}
\label{tab:pretraining_data}
\end{table*}

\begin{table}[t]
    \centering
    \scriptsize
    \begin{tabular}{ccccc}
\toprule
\textbf{Model} & \textbf{Text} & \textbf{Bbox} & \textbf{Size} & \textbf{Domain}\tabularnewline
\midrule
Donut & \checkmark &  & 13M & Synthetic, Doc\tabularnewline
Pix2Struct & \checkmark &  & 80M & Web\tabularnewline
QwenVL & \checkmark &  & 24.8M & Synthetic, Doc, Web\tabularnewline
UReader &  &  & 0.1M & Doc, Table, Chart, Web, Natural \tabularnewline
DocPedia & \checkmark &  & 0.9M & Doc \tabularnewline
CogAgent & \checkmark & \checkmark & 107M & Synthetic, Nature, Doc, Web\tabularnewline
DocOwl-1.5 & \checkmark & \checkmark & 4M & Doc, Table, Chart, Web, Natural\tabularnewline
\our{} & \checkmark & \checkmark & 357M & \makecell{Doc, Table, Chart, Web, Natural \\ Handwritten, Design, Math}\tabularnewline
\bottomrule
\end{tabular}

    \caption{Comparison of pre-training data used by document multimodal models.}
    \label{tab:comp_pretrain_data}
\end{table}

\section{\our{}}

\subsection{Model Architecture}
\label{subsection:model_architecture}
The architecture of \our{} comprises a vision encoder and a language decoder, connected through a resampling module to reduce the sequence length of the image~\cite{alayrac2022flamingo}, as illustrated in Figure~\ref{fig:model_arch}. The \textbf{vision encoder} is initialized from the Pix2Struct-Large model's encoder~\cite{lee2023pix2struct}, which is based on the Vision Transformer (ViT)~\cite{vit}. Consistent with Pix2Struct~\cite{lee2023pix2struct}, we employ a variable resolution strategy and extract the maximum number of fixed-size patches that can fit within a predefined sequence length. 

The \textbf{resampler} compresses the image sequence into a shorter, fixed number of tokens:
\begin{align}
    H_{0} = f(I)
\end{align}
\begin{align}
    H_{1} = \text{Attention}(V, [V; H_{0}], [V; H_{0}])
\end{align}
where $I$ is the input image, $f$ is the encoder function, $V$ represents a set of predefined soft tokens, and $[;]$ denotes the concatenation operator. The \textbf{language decoder} is based on a Transformer architecture and is designed to condition on both image and text contexts for next-token prediction. Details on the hyperparameters can be found in Appendix~\ref{supp:para}.

\subsection{Image and Text Representation}
The image representation is derived from the image encoder and resampler as described in Section~\ref{subsection:model_architecture}. Text representation is obtained through text tokenization and embedding. For markdown text, we directly tokenize it while preserving all special characters and formatting indicators. For text lines with bounding boxes, we convert the coordinates into discrete location tokens, similar to \textsc{Kosmos-2}~\cite{peng2023kosmos}.

We introduce a set of \(2L + 2\) specialized tokens: \texttt{<x$_0$>}, \texttt{<x$_1$>}, \ldots, \texttt{<x$_{L-1}$>}, \texttt{<y$_0$>}, \ldots, \texttt{<y$_{L-1}$>}, \texttt{<bbox>}, and \texttt{</bbox>}, which correspond to the coordinates and the start and end markers of a bounding box. The coordinates are obtained by rounding down the actual positions after resizing the images.

Consider a document \(T\) with \(N\) text lines. Each line is represented as \(\mathbf{T}_n = \{ w_1^{(n)}, w_2^{(n)}, \ldots, w_{M_n}^{(n)} \}\), where \(M_n\) is the number of words in the \(n\)-th text line. The bounding box for \(\mathbf{T}_n\) is then expressed as \(\mathbf{B}_n = \texttt{<bbox><} x_{\text{tl}}^{(n)} \texttt{><} y_{\text{tl}}^{(n)} \texttt{><} x_{\text{br}}^{(n)} \texttt{><} y_{\text{br}}^{(n)} \texttt{></bbox>}\), where the coordinates represent the top-left and bottom-right corners of the bounding box.

\subsection{Pre-training on Document Reading}

\subsubsection{Pre-training Tasks.}
Traditional Optical Character Recognition (OCR) tasks primarily focus on generating line-level text content and capturing its spatial positions within an image. While OCR preserves the layout positions of document text, it often overlooks the document-level reading order and structural integrity, both crucial for comprehensive document understanding. In contrast, markdown-formatted text provides an advantage over plain text by explicitly distinguishing various structural elements, such as tables and lists, using specific tokens. 

To effectively learn the layout and structure of documents, we propose two complementary generative tasks for pre-training a document foundation model: document-level text recognition and image-to-markdown generation, as detailed in Table~\ref{tab:task_prompts}.

\subsubsection{Training Objective and Formats.}
We train the model to predict outputs based on the input image context and task-specific prompts. The training objective is to minimize the cross-entropy loss for next-token prediction, commonly known as autoregressive language modeling~\cite{radford2018improving}. Table~\ref{tab:task_prompts} illustrates the formats for model training prompts and response sequences. 

The prompt is constructed by concatenating the image representation with a task-specific special token. The response corresponds to the text output of the tasks: text lines with bounding boxes for document-level text recognition and markdown text for the image-to-markdown generation task. A qualitative example is provided in Appendix~\ref{sp:qualitative_example} to illustrate the model’s input and output.

\subsubsection{Pre-training Data.}
Our training data is collected using an automated pipeline from diverse sources, resulting in a large corpus of 357.4 million document images, annotated with text lines using bounding boxes or in markdown format. As shown in Table~\ref{tab:pretraining_data}, our pre-training dataset encompasses a wide range of document types, including scanned documents, academic papers, web pages, design images, mathematical content, handwritten text, and more. Compared with the training data used by existing models in Table~\ref{tab:comp_pretrain_data}, \our{} leverages the largest and most diverse corpus, which significantly enhances the model’s adaptability and generalization across different domains.

We apply \textbf{filtering and quality control during data curation}. We use fastText for language identification (with a threshold of 0.5) to filter out non-English documents from the entire pre-training dataset. To ensure content diversity within each source, we use MinHash~\cite{broder1997resemblance} to identify and remove redundant pages, applying the same parameters as~\cite{lee2021deduplicating}, with document pairs having a similarity score of 0.8 or higher marked as duplicates. 

For image-to-markdown data sourced from README, DOCX, \LaTeX, and HTML files, we encountered discrepancies between the content in text images and their corresponding markdown sequences due to conversion issues. To refine the data, we evaluate token overlap between images and markdown files, requiring a token intersection-to-union ratio greater than 0.95 for inclusion. Details of the processing procedures for each document category are provided in Appendix~\ref{supp:dp}, along with sample training data in Appendix~\ref{supp:data}, aiming to offer transparent and reproducible guidelines for future research and applications.

\subsection{Fine-tuning on Document Understanding} 
\label{subsection:finetune}
We fine-tune \our{} on document understanding datasets, referring to the fine-tuned model as \ourchat{}. \ourchat{} is designed to answer diverse user-provided questions about text-intensive images from various domains. To better retain the reading capability of \our{}, we freeze the visual encoder of the pre-trained model and fine-tune the resampler and language model using a document understanding task prompt (Line 3 in Table~\ref{tab:task_prompts}), where [Question] and [Answer] represent a question-answer pair from the dataset.

\section{Experiments}

\subsection{Model and Training Configurations}
Following Pix2Struct~\cite{lee2023pix2struct}, we employ a short warmup phase of 20k steps to facilitate faster convergence during the pre-training stage. In this phase, the model learns to read text snippets from synthetic images rendered with random colors and fonts. Due to the substantially larger volume of layout-based data compared to markup-based data, we initially trained the model for 100k steps using only the layout-based dataset. We then combined the two datasets for an additional 140k steps of training. The total training involved approximately 260 billion tokens.

Our text tokenization is based on the \texttt{cl100k\_base} tiktoken tokenizer\footnote{\url{https://github.com/openai/tiktoken}}, with 8,194 specialized tokens introduced for coordinates and bounding box markers. The newly added word embeddings for location tokens are randomly initialized, with all parameters updated during training. We also incorporate data augmentation techniques from TrOCR~\cite{li2022trocr} to enhance the model's robustness.

\our{} contains a total of 1.3 billion parameters. Further details on model architecture and training hyperparameters are provided in Appendix~\ref{supp:para}.

\begin{table}[t]
    \small
    \centering
    \setlength{\tabcolsep}{1mm}
    \begin{tabular}{ccc}
    \toprule
    \bf Category & \bf Data Source & \bf Num \\
    \midrule
    \textbf{Handwritten} & Synthetic image & 200 \\
    \midrule
     \multirow{4}{*}{\bf Design} & MARIO-LAION~\cite{chen2024textdiffuser} & 200 \\
     & MARIO-OpenLibrary~\cite{chen2024textdiffuser} & 200 \\
     & MARIO-TMDB~\cite{chen2024textdiffuser} & 100 \\
    & MJ\&ST~\cite{gupta2016synthetic,jaderberg2014synthetic} & 200 \\
    \midrule
    \multirow{3}{*}{\textbf{Receipt}} &  Receipts crawled from the internet	& 100 \\
    & CORD \cite{park2019cord} & 100 \\
    & SROIE \cite{huang2019icdar2019} & 347 \\
    \midrule
    \textbf{Academic paper} & Academic papers from ArXiv & 200 \\	
    \midrule
    \multirow{3}{*}{\textbf{General}}& Financial statements from SEC	& 200 \\
    & General documents from Docx & 200 \\
    & FUNSD~\cite{jaume2019funsd} & 50 \\
    \midrule
    \textbf{Web Page} & Self-crawled web pages & 200 \\
    \midrule
    \multicolumn{2}{c}{\bf Total} & 2,297 \\
    \bottomrule
\end{tabular}

    \caption{Summary of document categories, data sources, and the number of samples in the OCREval benchmark.}
    \label{tab:summary_ocreval}
\end{table}

\begin{table}[t]
\small
    \centering
    \begin{tabular}{ccc}
    \toprule
    \bf Category & \bf Data Source & \bf Num \\
    \midrule
    \multirow{2}{*}{\textbf{Math Image}} &  CROHME Math & 1,000 \\
    & Ima2LaTeX-100k & 922 \\
    \midrule
    \textbf{Academic Paper} & ArXiv & 1,000 \\	
     \midrule
    \textbf{Table} & Table	& 771 \\
    \midrule
    \textbf{General Document} & Docx & 1,000 \\
    \midrule
    \textbf{Project Document} & README & 1,000 \\
    \midrule
    \multicolumn{2}{c}{\bf Total} & 5,693 \\
    \bottomrule
\end{tabular}

    \caption{Summary of document categories, data sources, and the number of samples in the MarkdownEval benchmark.}
    \label{tab:summary_markdowneval}
\end{table}

\begin{table*}[t]
    \scriptsize
    \centering
    \setlength{\tabcolsep}{1.2mm}
\begin{tabular}{ccccccccc}
\toprule
\bf Model & \bf Size & Handwritten & Design & Receipt & General & Academic & Web Image & Overall Score (Avg)\\
\midrule
Tesseract~\cite{TessOverview} & - & 42.3 / 58.1 / 62.6 & 24.3 / 26.2 / 26.4 & 63.7 / 49.3 / 65.1 & 92.1 / 56.8 / 86.7 & 78.1 / 56.9 / 91.7 & \textbf{90.4} / 54.0 / \textbf{75.5} & 65.2 / 50.2 / 68.0 \\
Nougat~\cite{blecher2023nougat} & 350M & 37.3 / - / 48.5 & 2.0 / - / 14.1 & 55.8 / - / 53.9 & 75.0 / - / 67.2 & 58.0 / - / 55.4 & 16.5 / - / 27.7 & 40.8 / - / 44.5 \\
Vary~\cite{Wei2023VarySU} & 7B & 28.0 / - / 62.4 & 43.1 / - / 75.8 & 31.8 / - / 62.4 & 55.9 / - / 54.2 & 45.6 / - / 49.4 & 10.1 / - / 26.4 & 35.8 / - / 55.1 \\
Qwen-VL~\cite{bai2023qwen} &9.6B & 53.6 / 70.8 / 74.8 & 7.6 / 28.3 / 29.2 & 43.0 / 37.7 / 48.0 & 76.6 / 78.1 / 74.6 & 52.2 / 65.5 / 58.4 & 19.8 / 37.5 / 34.1 & 42.1 / 53.0 / 53.2 \\
GPT-4o & - & 66.0 / 23.1 / 87.4 & \textbf{74.6} / 15.5 / \textbf{82.1} & 83.6 / 8.6 / 75.4 & 91.9 / 19.5 / 86.8 & 69.5 / 22.3 / 75.7 & 51.1 / 9.4 / 55.9 & 72.8 / 16.4 / 77.2 \\
KOSMOS-2.5 & 1.3B & \textbf{71.6} / \textbf{94.1} / \textbf{90.6} & 61.7 / \textbf{80.2} / 79.6 & \textbf{89.4} / \textbf{80.1} / \textbf{83.3} & \textbf{97.6} / \textbf{89.8} / \textbf{93.9} & \textbf{98.8} / \textbf{93.3} / \textbf{99.1} & 57.0 / \textbf{72.1} / 69.6 & \textbf{79.4} / \textbf{84.9} / \textbf{86.0} \\
\bottomrule
\end{tabular}

    \vspace{0.2cm}
    \caption{Experimental results for the document-level text recognition task on OCREval. Metrics are reported as F1$\uparrow$ / IOU$\uparrow$ / NED$\uparrow$. As Nougat and Vary produce only textual output without bounding boxes, IOU scores are not available for these models.}
    \label{tab:ocreval_results}
\end{table*}

\begin{table*}[h]
    \centering
    \scriptsize
    \begin{tabular}{cccccccc}
\toprule
\multirow{1}{*}{\bf Model}  & \bf Docx & \bf README & \bf Arxiv & \bf Tables & \bf Math Equation & \bf  CROHME Math & \bf Overall Score (Avg)\\
\midrule
MSOCR+T5\cite{Raffel2019T5} 
     & 73.1 / 6.7 & 72.8 / 4.2 & 55.2 / 4.6 & 32.4 / 13.0 & 13.3 / 0.9 & 30.3 / 5.4 
     & 46.2 / 5.8 \\
Nougat\cite{blecher2023nougat} 
     & 84.8 / 21.9 & 68.9 / 27.3 & 88.4 / 44.4 & 49.0 / 36.1 & 73.6 / 71.6 & 10.6 / 14.8 
     & 62.6 / 36.0 \\
Vary\cite{Wei2023VarySU} 
     & 85.4 / 46.3 & 72.5 / 35.6 & 80.6 / 70.2 & 29.3 / 25.2 & 30.4 / 44.7 & 11.5 / 34.2 
     & 51.6 / 42.7 \\
GPT-4o\footnote{\url{https://openai.com/index/hello-gpt-4o/}} 
     & 85.3 / 20.5 & 83.5 / 49.3 & 76.7 / 23.0 & 74.7 / 42.4 & 56.5 / 78.2 & 64.7 / 84.2 
     & 73.6 / 49.6 \\
\midrule
KOSMOS-2.5 
     & \bf 91.6 / 82.1 & \bf 95.1 / 91.2 & \bf 90.8 / 86.4 & \bf 85.1 / 90.1 & \bf 88.1 / 95.2 & \bf 98.5 / 99.7 
     & \bf 91.5 / 90.8 \\
\bottomrule
\end{tabular}

    \caption{Experimental results for document-level markdown generation on MDEval. Metrics are reported as NED$\uparrow$ / NTED$\uparrow$.}
    \label{tab:mdeval_results}
\end{table*}

\subsection{Evaluation on Document Reading}

\subsubsection{Benchmarks.}
To comprehensively evaluate models' capabilities in document-level text recognition and image-to-markdown generation tasks, we collected the OCREval and MarkdownEval benchmarks. The \textbf{OCREval benchmark} consists of 2,297 images from the test sets of 13 datasets, covering categories such as mathematical content, handwritten images, design images, receipts, digitally born documents, and web pages. The \textbf{MarkdownEval benchmark} includes 5,693 images spanning categories such as mathematical equations, academic papers, tables, general documents, and project documentation. The respective categories, data sources, and sample counts are detailed in Table~\ref{tab:summary_ocreval} and Table~\ref{tab:summary_markdowneval}. More data processing details are provided in Appendix~\ref{supp:eval_dp}.

\subsubsection{Metrics.}
The \textbf{metrics for OCREval} include word-level \textit{F1}, \textit{IOU}, and \textit{NED} to evaluate document-level OCR performance. The \textbf{metrics for MarkdownEval} include Normalized Edit Distance (NED) and Normalized Tree Edit Distance (NTED) for assessing image-to-markdown generation. NED is a string-based comparison metric, while NTED measures tree edit distance normalized by the number of nodes, capturing structural differences in parse trees. This dual evaluation framework considers both lexical accuracy and the preservation of the original hierarchical structure inherent in the Markdown format. Further details on the evaluation metrics are provided in Appendix~\ref{supp:ocr_evaluation_metrics} and Appendix~\ref{supp:evaluation_metrics}.

\paragraph{Results.}
\our{} is a unified framework that facilitates multitasking with tasks determined by the provided prompts. We compared \our{} against state-of-the-art document reading models on OCREval (Table~\ref{tab:ocreval_results}) and MarkdownEval (Table~\ref{tab:mdeval_results}). For the document-level text recognition task, \our{} outperforms existing models in reading text-intensive images. For instance, \our{} surpasses Vary$_{\text{Base}}$ by a significant margin despite having a smaller model size (1.3B vs. 7B parameters). \our{} also achieved the best performance across all image types on MarkdownEval. Notably, GPT-4o’s omission of markdown symbols affected its NTED scores slightly. For example, while \texttt{e}\^{} \texttt{2} should be represented as \texttt{e<sup>2</sup>} in markdown, GPT-4o outputs $e^2$ directly. For models adhering to markdown standards (e.g., Vary and Nougat), \our{} consistently outperforms them, benefiting from better layout understanding in text recognition.

\subsection{Evaluation on Document Understanding}

\paragraph{Settings.}
Fine-tuned on downstream datasets, \ourchat{} is capable of addressing a wide range of document understanding tasks. We fine-tuned \ourchat{} on the standard training sets of ten diverse document understanding datasets. These datasets cover general documents (DocVQA~\cite{mathew2020docvqa}, InfoVQA~\cite{infovqa}, DeepForm~\cite{deepform}, KLC~\cite{klc}), tables (WTQ~\cite{wikitableqa}, TabFact~\cite{TabFact}), charts (ChartVQA~\cite{chartqa}), natural images (TextVQA~\cite{textvqa}, TextCaps~\cite{textcaps}), and webpage screenshots (VisualMRC~\cite{visualmrc}). Evaluation is performed on the official test sets of nine public document understanding benchmarks. We did not evaluate on TextCaps due to the unavailability of the official evaluation server at this time.

\paragraph{Results.}
Table~\ref{tab:vqa_results} presents the experimental results compared to state-of-the-art OCR-free models. Among models with fewer than 2B parameters, \ourchat{} outperforms PixStruct$_{\text{LARGE}}$ and Donut across various benchmarks without task-specific fine-tuning. Compared to models exceeding 7B parameters, \ourchat{} delivers competitive performance on benchmarks covering documents, tables, and charts, including DocVQA, InfoVQA, DeepForm, KLC, WTQ, and ChartVQA. These results highlight the effectiveness of \ourchat{} in handling complex document understanding tasks.

\begin{table*}[t]
    \small
    \centering
        \begin{tabular}{ccccccccccc}
\toprule
\multirow{2}{*}{\textbf{Model}} & \multirow{2}{*}{\textbf{Size}} & \textbf{Doc} & \textbf{Info} & \textbf{Deep} & \multirow{2}{*}{\textbf{KLC}} & \multirow{2}{*}{\textbf{WTQ}} & \textbf{Tab} & \textbf{Chart} &  \textbf{Text} & \textbf{Visual}\tabularnewline 
 &  & \textbf{VQA} & \textbf{VQA} & \textbf{Form} &  &  & \textbf{Fact} & \textbf{QA} & \textbf{VQA} &  \textbf{MRC}\tabularnewline
\midrule
DocPeida~\cite{feng2023docpedia} & 7.0B & 47.1 & 15.2 & - & - & - & - & 46.9 & 60.2 & -\tabularnewline
DocOwl~\cite{mplug-docowl} & 7.1B & 62.2 & 38.2 & 42.6 & 30.3 & 26.9 & 60.2 & 57.4 & 52.6 & 188.8\tabularnewline
QwenVL~\cite{bai2023qwen} & 9.6B & 65.1 & 35.4 & - & - & - & - & 65.7 & 63.8 & -\tabularnewline
UReader~\cite{ureader} & 7.1B & 65.4 & 42.2 & 49.5 & 32.8 & 29.4 & 67.6 & 59.3 & 57.6 & 221.7\tabularnewline
Monkey~\cite{li2023monkey} & 9.8B & 66.5 & 36.1 & 40.6 & 32.8 & 25.3 & - & - & 67.6 & -\tabularnewline
HRVDA~\cite{liu2024hrvda} & 7.1B & 72.1 & 43.5 & 63.2 & 37.5 & 31.2 & 72.3 & 67.6 & 73.3 & 211.5\tabularnewline
DocOwl-1.5~\cite{mplug-docowl-1.5} & 8.1B & 81.6 & 50.4 & 68.8 & 37.9 & 39.8 & 80.4 & 70.5 & 68.8 & 239.5\tabularnewline
CogAgent~\cite{hong2024cogagent} & 17.3B & 81.6 & 44.5 & - & - & - & - & 68.4 & 76.1 & -\tabularnewline
\midrule
Donut$^{*}$~\cite{kim2021donut} & $<$1B & 67.5 & 11.6 & 61.6 & 30.0 & 18.8 & \textbf{54.6} & 41.8 & 43.5 & 93.9\tabularnewline
Dessurt$^{*}$~\cite{dessurt} & $<$1B & 63.2 & - & - & - & - & - & - & - & -\tabularnewline
Pix2Struct$^{*}_{\text{LARGE}}$~\cite{lee2023pix2struct} & 1.3B & 76.6 & 40.0 & - & - & - & - & 58.6 & - & -\tabularnewline
Vary-toy~\cite{varytoy} & 1.8B & 65.6 & - & - & - & - & - & 59.1 & - & -\tabularnewline
MiniCPM-V 2.0~\cite{yao2024minicpmvgpt4vlevelmllm} & 2.8B & 71.9 & - & - & - & - & - & 55.6 & \textbf{74.1} &  -\tabularnewline

\textsc{Kosmos-2.5-chat} & 1.3B & \textbf{81.1} & \textbf{41.3} & \textbf{65.8} & \textbf{35.1} & \textbf{32.4} & 49.9 & \textbf{62.3} & 40.7 & \textbf{156.0}\tabularnewline
\bottomrule
\end{tabular}

    \caption{Experimental results on document understanding benchmarks. The models listed above the line have more than 7B parameters, while those below the line are smaller models. The superscript ‘$*$’ indicates models fine-tuned separately on each downstream task. Among models with fewer than 7B parameters, the best results are marked in \textbf{bold}.}
    \label{tab:vqa_results}
\end{table*}

\section{Related Work}

\subsection{Multimodal Large Language Models}

Multimodal large language models (MLLMs) can be broadly categorized into LLM-centric scheduling systems and end-to-end trainable multimodal systems. LLM-centric scheduling systems leverage various vision foundation models, orchestrating them in a language-centric manner~\cite{wu2023visual,yang2023mm,liang2023taskmatrix,shen2023hugginggpt,liu2023internchat,suris2023vipergpt,chen2023language}. On the other hand, end-to-end trainable multimodal systems integrate vision and language models into a unified framework~\cite{metalm,alayrac2022flamingo,huang2023language,peng2023kosmos,huang2021seeing,xue2021probing,zhu2023minigpt,huang2023sparkles,li2023blip,dai2023instructblip,liu2023visual,luo2023cheap,wang2023visionllm,su2023pandagpt,zhang2023llama,gao2023llama,koh2023grounding,li2023otter,tang2024any,tang2023codi,kondratyuk2023videopoet,bai2023qwen,hong2024cogagent,yao2024minicpmvgpt4vlevelmllm,varytoy}. 

Our model falls into the latter category, sharing similarities with grounded multimodal models like KOSMOS-2~\cite{peng2023kosmos}, Shikra~\cite{chen2023shikra}, and ChatSpot~\cite{zhao2023chatspot}, which output object locations in natural images. However, \our{} uniquely focuses on text-image reading and understanding capabilities, tackling the challenge of producing high-quality document layouts while maintaining the structural integrity crucial for document understanding.

\subsection{Document Reading and Understanding}

Document reading and understanding leverage AI to automatically read, comprehend, and extract information from documents~\cite{cui2021document,xu2020layoutlm,xu-etal-2021-layoutlmv2,xu2021layoutxlm,huang2022layoutlmv3,kim2021donut,chen2022xdoc,li2021markuplm,li2022dit,li2021selfdoc,appalaraju2021docformer,wang2022lilt,gu2022xylayoutlm,li2021structurallm,chen2024textdiffuser,yu2023structextv2,li2023monkey,liu2024textmonkey}. Representative document foundation models like LayoutLMv3 integrate text, layout, and image information during pre-training, excelling in tasks like key information extraction and document question answering~\cite{huang2022layoutlmv3}. Donut~\cite{kim2021donut} introduces an OCR-free document understanding Transformer, directly mapping input document images to desired outputs. Models like Pix2Struct~\cite{lee2023pix2struct}, HRVDA~\cite{liu2024hrvda}, and the mPLUG-DocOwl series~\cite{mplug-docowl,mplug-docowl-1.5} pre-train vision encoders on document reading tasks, resulting in impressive document understanding performance. \our{} scales up document pre-training to include up to 357.4 million document pages and more challenging tasks, significantly enhancing the model's reading and understanding capabilities.

Nougat~\cite{blecher2023nougat} similarly parses documents into markup language, but its focus is limited to scientific documents. In contrast, \our{} excels across a broader range of documents and generalizes well to document understanding tasks.
Recent works like DocPedia~\cite{feng2023docpedia} enhance MLLMs’ text-rich image understanding by processing visual input in the frequency domain for high-resolution capabilities.
Approaches like TextSquare~\cite{tang2024textsquare}, TRINS~\cite{zhang2024trins}, and LLaVAR~\cite{zhang2023llavar} enhance reading abilities by using publicly available OCR tools and closed-source MLLMs to generate instruction-tuning data for text-rich images. LLaVA-read further uses open-source OCR tools to extract text and layout information for language models. UReader~\cite{ureader} introduces a shape-adaptive cropping module to efficiently encode low-resolution sub-images. Meanwhile, Monkey~\cite{li2023monkey} boosts training efficiency and resolution, excelling in image captioning and text-rich document processing. However, these methods rely on pre-trained vision encoders without document-specific pre-training, which limits their performance. 
After extensive pre-training, \our{} achieves strong document understanding performance by fine-tuning on publicly available benchmarks only, without needing complex module designs, OCR tools, or closed-source MLLMs.

\subsection{Document Reading Benchmarks}

Existing OCR evaluation benchmarks like OCRBench~\cite{Liu2023ocrbench} or DocLocal4K~\cite{mplug-docowl-1.5} mainly focus on text-line recognition tasks. Textmonkey~\cite{liu2024textmonkey} evaluates the model on natural images only. In contrast, our proposed OCREval is the first benchmark specifically designed to assess document-level text recognition, which demands more advanced recognition capabilities. For markdown evaluation, 
Nougat~\cite{blecher2023nougat} restricts its performance assessment to academic papers from ArXiv. In contrast, our MarkdownEval offers a more comprehensive assessment by covering a wider range of image domains, providing a more robust assessment of model capabilities.
\section{Conclusion and Future Work}


In summary, this work advances document-level machine reading by introducing a novel pre-training framework and demonstrating its effectiveness through impressive performance on diverse benchmarks.
Our pre-trained model, \our{}, excels in document reading, while our fine-tuned model, \ourchat{}, achieves competitive results in document understanding benchmarks.
The extensive corpus of 357.4 million annotated document images and the development of OCREval and MarkdownEval benchmarks provide comprehensive tools for evaluating and furthering research in document intelligence.
Despite these promising results, our current model faces some limitations, offering valuable future research directions. 
For instance, documents spanning multiple pages pose a challenge as they typically demand holistic processing and comprehension. Meanwhile, it is also feasible that \our{} allows for multiple image pages interleaved with text as input; however, managing long context remains a vital issue we aim to address in future work.
In the broader research landscape, a significant direction lies in advancing model scaling capabilities. With an expanding range of tasks and complexities, scaling the model to handle larger data volumes is crucial for multimodal literate models.

\bibliography{aaai25}

\clearpage
\newpage

\appendix
\section{Appendix}

\subsection{Model and Training Hyperparameters}
\label{supp:para}

Model and training hyperparameters are demonstrated in Table~\ref{tab:basic_hyperparameters} and Table~\ref{tab:training_hyperparameters}.

\begin{table}[ht]
\small
\centering
\begin{tabular}{ccc}
\toprule
\textbf{Modules} & \textbf{Hyperparameters} & \\
\midrule
\multirow{8}{*}{\textbf{Image Encoder}} & Patch size & 16 \\
& Patch embed hidden size & 768 \\
& Number of layers & 18 \\
& Hidden size & 1,536 \\
& FFN inner hidden size & 3,968 \\
& Attention heads & 24 \\
& Activation function & GeLU \\
& Max sequence length & 4,096 \\
\midrule
\multirow{3}{*}{\textbf{Resampler}} & Number of layers & 1 \\
& Hidden size & 1,536 \\
& Output sequence length & 2,048 \\
\midrule
\multirow{9}{*}{\textbf{Language Decoder}} & Number of layers & 24 \\
& Hidden size & 1,536 \\
& FFN inner hidden size & 6,144 \\
& Attention heads & 16 \\
& Activation function & GeLU \\
& Vocabulary size & 108,481 \\

& Max sequence length & 4,096 \\
\bottomrule
\end{tabular}

\caption {Model Hyperparameters of \our{}}
\label{tab:basic_hyperparameters}
\end{table}

\begin{table}[ht]
\small
\centering
\begin{tabular}{lcc}
\toprule
\textbf{Hyperparameters} & \textbf{Pre-training} & \textbf{Fine-tuning}\\
\midrule
Training steps & 260,000 &  3000\\
Warmup steps & 375 &  100\\
Batch size & \multicolumn{2}{c}{1,024}\\
Optimizer & \multicolumn{2}{c}{AdamW}  \\
Learning rate & \multicolumn{2}{c}{2e-4} \\
Learning rate decay & \multicolumn{2}{c}{Linear} \\
Adam $\beta$ & \multicolumn{2}{c}{(0.9, 0.98)} \\
Weight decay & \multicolumn{2}{c}{0.01} \\
Dropout & \multicolumn{2}{c}{0.1} \\
\bottomrule
\end{tabular}

\caption{Training hyperparameters of \our{}}
\label{tab:training_hyperparameters}
\end{table}

\subsection{OCR Evaluation Metrics}
\label{supp:ocr_evaluation_metrics}
\paragraph{F1. }
The F1 score is a commonly used evaluation metric for measuring the accuracy of models in classification tasks. It is the harmonic mean of Precision and Recall, offering a balanced measure that considers both false positives and false negatives. In the OCR task, the F1 score will be used to assess the effectiveness of OCR models in recognizing words from images.
Precision is the ratio of correctly recognized words to the total number of words detected by the model. Recall is the ratio of correctly recognized words to the total number of actual words. The F1 score is the harmonic mean of Precision and Recall, and it is calculated as follows:

\begin{equation*}
    \text{Precision} = \frac{\text{TP}}{\text{TP} + \text{FP}}
\end{equation*}
\begin{equation*}
    \text{Recall} = \frac{\text{TP}}{\text{TP} + \text{FN}}
\end{equation*}
\begin{equation*}
    F1 = 2 \cdot \frac{\text{Precision} \cdot \text{Recall}}{\text{Precision} + \text{Recall}}
\end{equation*}

where TP is the number of correctly recognized words, FP is the number of incorrectly recognized words, and FN is the number of missed words.

\paragraph{IoU. }
Intersection over Union (IoU) is a critical evaluation metric for assessing the performance of object detection models, including OCR textline detection. IoU measures the overlap between the predicted bounding box and the ground truth bounding box, providing a quantitative measure of how well the model has detected the textlines in an image. The formula for IoU is as follows:
\begin{equation*}
    \text{IoU} = \frac{\text{Area of Intersection}}{\text{Area of Union}}
\end{equation*}

\paragraph{NED.}
Normalized Edit Distance (NED) is an extension of the Edit Distance (Levenshtein Distance) metric, commonly used to assess the similarity between two text strings. The calculation of NED can be found in Appendix~\ref{supp:evaluation_metrics}.

\subsection{Image-to-markdown Generation Evaluation Metrics}
\label{supp:evaluation_metrics}
In light of the unique nature of the image-to-markdown conversion task, assessing the quality of the generated markdown necessitates specialized metrics. We adopt a two-fold evaluation scheme: Normalized Edit Distance (NED) and Normalized Tree Edit Distance (NTED), considering both the lexical accuracy and the preservation of the original structural elements.
The NED is formulated as 
\begin{equation*}
    \textit{NED} = 1-\frac{1}{N} \sum_{i=1}^N D\left(s_i, \hat{s}_i\right) / \max \left(\mathrm{len}(s_i), \mathrm{len}(\hat{s}_i\right))
\end{equation*}
where $N$, $s$, and $\hat{s}$ denote the number of samples, prediction, and ground truth, respectively. $D(\cdot,\cdot)$ and $\mathrm{len}(\cdot)$ represent the edit distance function and the length of a string. The \textit{NED} value ranges from 0 to 1, with a higher \textit{NED} value indicating the prediction is closer to the ground truth.

However, given the hierarchical structure inherent to markdown, relying solely on a string-based comparison metric like NED can be insufficient. Thus, we adopt NTED as an additional evaluation metric for structural differences. NTED is a tree edit distance normalized by the number of nodes in the tree, considering the structural discrepancies between parse trees. Specifically, the predicted markdown sequence is first transformed into an HTML tree. Then, the tree edit distance between the prediction and the ground truth is calculated using the ZSS algorithm \cite{zhang1989simple}. The NTED is formulated as
\begin{equation*}
\textit{NTED} = 1-\frac{1}{N} \sum_{i=1}^N \mathrm{TD}\left(t_i, \hat{t}_i\right) / \max \left(\mathrm{node}(t_i), \mathrm{node}(\hat{t}_i\right))
\end{equation*}
where $N$, $t$, and $\hat{t}$ signify the number of samples, the HTML tree of prediction, and the HTML tree of ground truth, respectively. Besides, $\mathrm{TD}(\cdot,\cdot)$ and $\mathrm{node}(\cdot)$ stand for the tree edit distance function and the number of nodes in a tree.

\subsection{Qualitative Example}
\label{sp:qualitative_example}
\begin{figure*}[t]
  \centering
  \begin{subfigure}{0.31\textwidth}
    \centering
    \frame{\includegraphics[width=\textwidth]{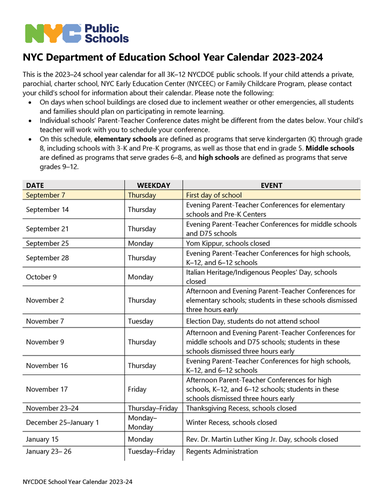}}
    \caption{Input}
    \label{fig:models_in}
  \end{subfigure}%
  ~
  \begin{subfigure}{0.31\textwidth}
    \centering
    \frame{\includegraphics[width=\textwidth]{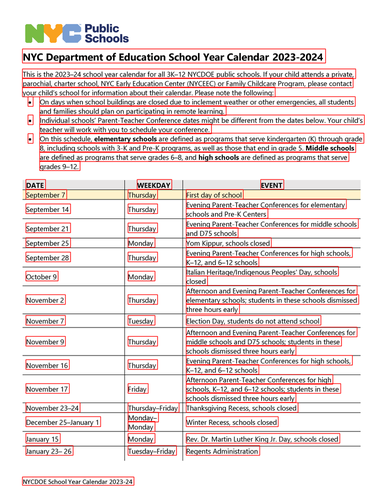}}
    \caption{Using the layout prompt}
    \label{fig:models_ocr}
  \end{subfigure}
  ~
  \begin{subfigure}{0.31\textwidth}
    \centering
    \frame{\includegraphics[width=\textwidth]{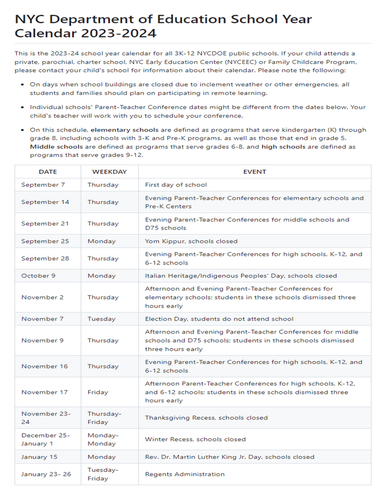}}
    \caption{Using the markup prompt}
    \label{fig:models_md}
  \end{subfigure}

  \caption{\our{}'s outputs given the same text image and different task prompts.}
  \label{fig:exp}
\end{figure*}

We illustrate an example in Figure~\ref{fig:exp}, showcasing the model outputs produced by \our{} with various task prompts when presented with the same input text image. 
For document-level text recognition task, \our{} produces the following text sequence, which includes textual content and corresponding bounding boxes:
\vspace{2mm}
\begin{lstlisting}[backgroundcolor = \color{gray!5}, basicstyle=\tiny\ttfamily, mathescape]
[x_52] [y_113] [x_756] [y_145]: NYC Department of Education School Year Calendar 2023-2024
[x_52] [y_159] [x_826] [y_181]: This is the 2023-24 school year calendar for all 3K-12 NYCDOE public schools. If your child attends a private,
[x_52] [y_180] [x_820] [y_202]: parochial, charter school, NYC Early Education Center (NYCEEC) or Family Childcare Program, please contact
[x_52] [y_201] [x_639] [y_223]: your child's school for information about their calendar. Please note the following:
[x_65] [y_223] [x_77] [y_245]: $\bullet$
[x_92] [y_223] [x_825] [y_245]: On days when school buildings are closed due to inclement weather or other emergencies, all students
... 
\end{lstlisting}
\vspace{2mm}
For image-to-markdown generation task, \our{} generates the text sequence in Markdown format:
\vspace{2mm}
\begin{lstlisting}[backgroundcolor = \color{gray!5}, basicstyle=\tiny\ttfamily, mathescape]
# NYC Department of Education School Year Calendar 2023-2024

This is the 2023-24 school year calendar for all 3K-12 NYCDOE public schools. If your child attends a private, parochial, charter school, NYC Early Education Center (NYCEEC) or Family Childcare Program, please contact your child's school for information about their calendar. Please note the following:
... 
-   On this schedule, **elementary schools** are defined as programs that serve kindergarten (K) through grade 8, including schools with 3-K and Pre-K programs, as well as those that end in grade 5. **Middle schools** are defined as programs that serve grades 6-8, and **high schools** are defined as programs that serve grades 9-12.
...
\end{lstlisting}
\vspace{2mm}
The example shows that \our{} precisely identifies text positions and recognizes text content. Moreover, it adeptly captures the styles and structures present within the text image, including elements like titles, bullet points, tables, and bold text. Section~\ref{supp:example} provides the full output sequence using different task prompts for this example. Furthermore, \our{} is compatible with more powerful LLMs like GPT-3.5 or GPT-4. The output from our model can serve as contexts for LLMs, enhancing their capabilities through further prompt engineering.

\subsection{Pre-training Data Processing}
\label{supp:dp}

The pre-training data has a wide coverage, and each type of data requires a different processing workflow, which is introduced as follows:

\paragraph{Scanned Document} We use the Microsoft Read API \footnote{\url{https://learn.microsoft.com/en-us/azure/ai-services/computer-vision/overview-ocr\#read-api}} to extract text and layout information.

\paragraph{General Document, Academic Paper, Design} We first compile and convert arXiv papers, SEC files, and PowerPoint slides into PDF files. Together with other general PDFs and poster, we employed the PyMuPDF parser \footnote{\url{https://github.com/pymupdf/PyMuPDF}} to extract text and layout information efficiently.

\paragraph{Web Page} We also include webpage screenshots in the model pre-training to diversify the layout distribution further. We collect the webpage URLs from the English portion of the mC4 dataset. Playwright \footnote{\url{https://github.com/microsoft/playwright-python}} is used to access a specified URL and open the webpage. The HTML content of the page is extracted and parsed using the lxml library \footnote{\url{https://lxml.de/}} to obtain a Document Object Model (DOM) tree representation. This DOM tree is traversed, examining the XPath of each element within it. This traversal aims to determine whether each element is visible and retrieve information about its bounding boxes.

\paragraph{Mathematical} In CROHME, each piece of data is an individual formula. We randomly select between 5 to 15 formulas and paste them onto a single blank page at random positions for formula recognition in page level.

\paragraph{Handwritten} We downloaded 5,427 handwritten fonts from Google Fonts, and for each textline generation, we randomly select one of these fonts.

\paragraph{General Document (markdown)} The Microsoft Office WORD files have been extensively used in existing research like TableBank~\citep{li2020tablebank} and ReadingBank~\citep{wang2021layoutreader}. We collect WORD DOCX files and convert them into texts with markdown. First, we use Pandoc to convert the XML content within the DOCX files into markdown files. As Pandoc keeps the ``<table>'' tags to represent the tabular cells in the generated markdown, we further identify all the tables and use markdownify \footnote{\url{https://github.com/matthewwithanm/python-markdownify}} to convert them into the markdown formats. Finally, the original DOCX files are converted into PDF files, and each page is aligned to the corresponding span of the markdown content based on a heuristic method. 

\paragraph{Academic Paper (markdown)} \LaTeX\ documents from arXiv have been used to generate PDF files to obtain texts with bounding boxes. Meanwhile, we also convert the \LaTeX\ content into the markdown texts. Similar to Nougat~\citep{blecher2023nougat}, LaTeXML \footnote{\url{https://math.nist.gov/~BMiller/LaTeXML/}} is used to convert the \LaTeX\ code into the HTML sequence, which is further transformed into the markdown format. Different from Nougat, we keep all the tables at the beginning of the page as most \LaTeX\ users prefer to position tables with ``[t]'' or ``[h]'' instead of ``[b]''. Meanwhile, we also convert the table content from the \LaTeX\ format into the markdown format.

\paragraph{Project Document (markdown)} In addition to layout-based data, we collect markup-based data for the pre-training. We collect ``README.md'' files from many GitHub projects and convert these files into HTML using Pandoc \footnote{\url{https://pandoc.org/}}. Then, wkhtmltopdf \footnote{\url{https://wkhtmltopdf.org/}} is used to obtain the images from the generated HTML content. 

\paragraph{Web Page (markdown)} The most straightforward way to obtain markdown resources from HTML webpages is through web scraping. However, webpages are often cluttered with various layouts and styles, resulting from the misuse of HTML tags. Moreover, HTML pages may include extraneous elements, such as advertisements, navigation menus, or formatting elements, making extracting clean and meaningful content challenging. To overcome these obstacles, we employ Playwright, a fast and reliable end-to-end testing framework for the web. The library allows us to navigate the HTML structure, filter out non-essential elements, and extract the relevant text content. We also apply custom rules and regular expressions to further refine the extracted text and format it as markdown, ensuring that the resulting markdown files are coherent and readable.

\subsection{Evaluation Data Processing}
\label{supp:eval_dp}

\paragraph{OCREval. } We constructed the OCREval comprising 2,297 samples, covering data from various domains. The details of each dataset's construction are provided below:
\begin{itemize}
\item \textbf{Design. }
We used the Microsoft Read API to obtain OCR results for MARIO-LAION, MARIO-OpenLibrary, and MARIO-TMDB, followed by manual verification to ensure the accuracy of the ground truth. For MJ\&ST, MJSynth consists of single-line textlines, which we randomly selected and placed multiple textlines on a single page to create page-level OCR test samples. For SynthText, we used the provided text and bounding box as the OCR ground truth.
\item \textbf{Receipt. }
For SROIE and CORD, we use their official annotations, which are carefully annotated by crowd workers and not from third-party OCR results. For Receipts crawled from the internet, we used Bing's image search engine\footnote{\url{https://www.bing.com/images}} with the keyword "receipt" to find relevant images. Subsequently, we used the Microsoft Read API to obtain OCR results and manually verified them, filtering out non-English receipts.

\item \textbf{Others. } 
The processing steps for Handwritten, Academic paper, General, and Web Page are consistent with those used in the pre-training phase, as detailed in Appendix~\ref{supp:dp}.
\end{itemize}

To ensure the accuracy of the OCR ground truth, we utilized the provided OCR ground truth for publicly available datasets. For datasets without provided ground truth and for our own constructed dataset, we obtained OCR results using Microsoft's Read OCR engine\footnote{\url{https://learn.microsoft.com/en-us/azure/ai-services/computer-vision/overview-ocr}} and then manually check them to ensure accuracy.

\paragraph{MarkdownEval. }
We constructed a dataset called markdownEval, consisting of 5,633 test samples, to evaluate the model's understanding of image across various domains. The details of each dataset's construction are provided below:

\begin{itemize}
\item \textbf{Math Image. }
Both CROHME Math and Ima2LaTeX-100k consist of formulas and their corresponding LaTeX source code. We used Pandoc\footnote{\url{https://pandoc.org/}} to convert the LaTeX source code into Markdown format and then randomly selected multiple samples to place on a single page to create test samples.

\item \textbf{Table. }
We extracted the LaTeX source code for tables from arXiv sources and then compiled it using pdfLaTeX\footnote{\url{https://www.tug.org/texlive/}} to obtain table images. Subsequently, we used Pandoc to convert the LaTeX source code into Markdown format to create test samples.

\item \textbf{Others. }
The processing steps for Project Document, Academic Paper, and General Document are consistent with those used in the pre-training phase, as detailed in Appendix~\ref{supp:dp}.

\end{itemize}

\subsection{Examples of Model Inference}
\label{supp:example}

\begin{lstlisting}[backgroundcolor = \color{gray!5}, basicstyle=\ttfamily, basicstyle=\tiny, mathescape, caption=Model outputs using the layout-based prompt]
[x_52] [y_113] [x_756] [y_145]: NYC Department of Education School Year Calendar 2023-2024
[x_52] [y_159] [x_826] [y_181]: This is the 2023-24 school year calendar for all 3K-12 NYCDOE public schools. If your child attends a private,
[x_52] [y_180] [x_820] [y_202]: parochial, charter school, NYC Early Education Center (NYCEEC) or Family Childcare Program, please contact
[x_52] [y_201] [x_639] [y_223]: your child's school for information about their calendar. Please note the following:
[x_65] [y_223] [x_77] [y_245]: $\bullet$
[x_92] [y_223] [x_825] [y_245]: On days when school buildings are closed due to inclement weather or other emergencies, all students
[x_92] [y_244] [x_525] [y_266]: and families should plan on participating in remote learning.
[x_65] [y_265] [x_77] [y_287]: $\bullet$
[x_92] [y_265] [x_846] [y_287]: Individual schools' Parent-Teacher Conference dates might be different from the dates below. Your child's
[x_92] [y_286] [x_491] [y_308]: teacher will work with you to schedule your conference.
[x_65] [y_308] [x_77] [y_330]: $\bullet$
[x_92] [y_307] [x_845] [y_330]: On this schedule, elementary schools are defined as programs that serve kindergarten (K) through grade
[x_92] [y_329] [x_826] [y_351]: 8, including schools with 3-K and Pre-K programs, as well as those that end in grade 5. Middle schools
[x_92] [y_350] [x_810] [y_372]: are defined as programs that serve grades 6-8, and high schools are defined as programs that serve
[x_92] [y_371] [x_186] [y_393]: grades 9-12.
[x_60] [y_414] [x_106] [y_436]: DATE
[x_318] [y_414] [x_399] [y_436]: WEEKDAY
[x_605] [y_414] [x_659] [y_436]: EVENT
[x_60] [y_437] [x_155] [y_459]: September 7
[x_297] [y_437] [x_366] [y_459]: Thursday
[x_432] [y_437] [x_565] [y_459]: First day of school
[x_60] [y_470] [x_164] [y_492]: September 14
[x_297] [y_470] [x_366] [y_492]: Thursday
[x_432] [y_459] [x_804] [y_481]: Evening Parent-Teacher Conferences for elementary
[x_432] [y_480] [x_622] [y_503]: schools and Pre-K Centers
[x_60] [y_514] [x_164] [y_536]: September 21
[x_297] [y_514] [x_366] [y_536]: Thursday
[x_432] [y_504] [x_832] [y_526]: Evening Parent-Teacher Conferences for middle schools
[x_432] [y_525] [x_553] [y_547]: and D75 schools
[x_60] [y_548] [x_164] [y_570]: September 25
[x_297] [y_548] [x_360] [y_570]: Monday
[x_432] [y_548] [x_630] [y_570]: Yom Kippur, schools closed
[x_60] [y_581] [x_164] [y_603]: September 28
[x_297] [y_581] [x_366] [y_603]: Thursday
[x_432] [y_570] [x_818] [y_593]: Evening Parent-Teacher Conferences for high schools,
[x_432] [y_592] [x_601] [y_614]: K-12, and 6-12 schools
[x_60] [y_625] [x_135] [y_647]: October 9
[x_297] [y_625] [x_360] [y_647]: Monday
[x_432] [y_614] [x_786] [y_636]: Italian Heritage/Indigenous Peoples' Day, schools
[x_432] [y_636] [x_482] [y_658]: closed
[x_60] [y_679] [x_152] [y_701]: November 2
[x_297] [y_679] [x_366] [y_701]: Thursday
[x_432] [y_658] [x_829] [y_680]: Afternoon and Evening Parent-Teacher Conferences for
[x_432] [y_679] [x_833] [y_701]: elementary schools; students in these schools dismissed
[x_432] [y_700] [x_556] [y_723]: three hours early
[x_60] [y_727] [x_152] [y_749]: November 7
[x_297] [y_727] [x_360] [y_749]: Tuesday
[x_432] [y_727] [x_745] [y_749]: Election Day, students do not attend school
[x_60] [y_775] [x_152] [y_797]: November 9
[x_297] [y_775] [x_366] [y_797]: Thursday
[x_432] [y_754] [x_829] [y_776]: Afternoon and Evening Parent-Teacher Conferences for
[x_432] [y_775] [x_793] [y_797]: middle schools and D75 schools; students in these
[x_432] [y_796] [x_687] [y_818]: schools dismissed three hours early
[x_60] [y_829] [x_161] [y_851]: November 16
[x_297] [y_829] [x_366] [y_851]: Thursday
[x_432] [y_819] [x_818] [y_841]: Evening Parent-Teacher Conferences for high schools,
[x_432] [y_840] [x_601] [y_862]: K-12, and 6-12 schools
[x_60] [y_884] [x_161] [y_906]: November 17
[x_297] [y_884] [x_344] [y_906]: Friday
[x_432] [y_863] [x_773] [y_885]: Afternoon Parent-Teacher Conferences for high
[x_432] [y_884] [x_791] [y_906]: schools, K-12, and 6-12 schools; students in these
[x_432] [y_905] [x_687] [y_927]: schools dismissed three hours early
[x_60] [y_928] [x_186] [y_950]: November 23-24
[x_297] [y_928] [x_416] [y_950]: Thursday-Friday
[x_432] [y_928] [x_692] [y_950]: Thanksgiving Recess, schools closed
[x_60] [y_960] [x_234] [y_983]: December 25-January 1
[x_297] [y_950] [x_368] [y_972]: Monday-
[x_297] [y_971] [x_360] [y_994]: Monday
[x_432] [y_960] [x_646] [y_983]: Winter Recess, schools closed
[x_60] [y_999] [x_140] [y_1021]: January 15
[x_297] [y_999] [x_360] [y_1021]: Monday
[x_432] [y_999] [x_789] [y_1021]: Rev. Dr. Martin Luther King Jr. Day, schools closed
[x_60] [y_1027] [x_170] [y_1049]: January 23- 26
[x_297] [y_1027] [x_410] [y_1049]: Tuesday-Friday
[x_432] [y_1027] [x_603] [y_1049]: Regents Administration
[x_52] [y_1099] [x_311] [y_1118]: NYCDOE School Year Calendar 2023-24
\end{lstlisting}

\begin{lstlisting}[backgroundcolor = \color{gray!5}, basicstyle=\ttfamily, basicstyle=\tiny, mathescape, caption=Model outputs using the markup-based prompt]
# NYC Department of Education School Year Calendar 2023-2024

This is the 2023-24 school year calendar for all 3K-12 NYCDOE public schools. If your child attends a private, parochial, charter school, NYC Early Education Center (NYCEEC) or Family Childcare Program, please contact your child's school for information about their calendar. Please note the following:

-   On days when school buildings are closed due to inclement weather or other emergencies, all students and families should plan on participating in remote learning.

-   Individual schools' Parent-Teacher Conference dates might be different from the dates below. Your child's teacher will work with you to schedule your conference.

-   On this schedule, **elementary schools** are defined as programs that serve kindergarten (K) through grade 8, including schools with 3-K and Pre-K programs, as well as those that end in grade 5. **Middle schools** are defined as programs that serve grades 6-8, and **high schools** are defined as programs that serve grades 9-12.

| DATE | WEEKDAY | EVENT |
| --- | --- | --- |
| September 7 | Thursday | First day of school |
| September 14 | Thursday | Evening Parent-Teacher Conferences for elementary schools and Pre-K Centers |
| September 21 | Thursday | Evening Parent-Teacher Conferences for middle schools and D75 schools |
| September 25 | Monday | Yom Kippur, schools closed |
| September 28 | Thursday | Evening Parent-Teacher Conferences for high schools, K-12, and 6-12 schools |
| October 9 | Monday | Italian Heritage/Indigenous Peoples' Day, schools closed |
| November 2 | Thursday | Afternoon and Evening Parent-Teacher Conferences for elementary schools; students in these schools dismissed three hours early |
| November 7 | Tuesday | Election Day, students do not attend school |
| November 9 | Thursday | Afternoon and Evening Parent-Teacher Conferences for middle schools and D75 schools; students in these schools dismissed three hours early |
| November 16 | Thursday | Evening Parent-Teacher Conferences for high schools, K-12, and 6-12 schools |
| November 17 | Friday | Afternoon Parent-Teacher Conferences for high schools, K-12, and 6-12 schools; students in these schools dismissed three hours early |
| November 23-24 | Thursday-Friday | Thanksgiving Recess, schools closed |
| December 25-January 1 | Monday- Monday | Winter Recess, schools closed |
| January 15 | Monday | Rev. Dr. Martin Luther King Jr. Day, schools closed |
| January 23- 26 | Tuesday-Friday | Regents Administration |
\end{lstlisting}

\subsection{Pre-training Data Examples}
\label{supp:data}

We demonstrate some of the pre-training data examples used in \our{}, which include the input and output from IIT-CDIP, general pdfs, SEC, arXiv papers, web screenshots, PowerPoint slides, poster, mathematical, handwrittens, README, DOCX, \LaTeX\, and HTML.

\begin{figure*}[htbp]
  \centering

  \begin{subfigure}{0.4\textwidth}
    \centering
    \frame{\includegraphics[width=\textwidth]{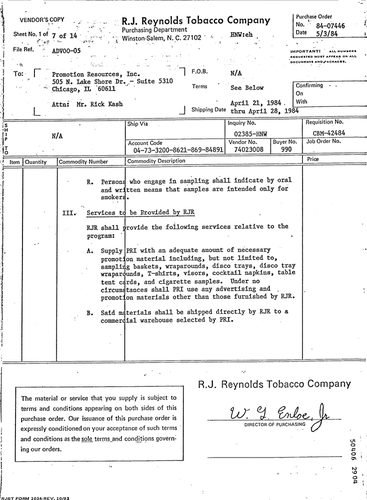}}
    \caption{Input}
    \label{fig:cdip_sub1}
  \end{subfigure}%
  ~
  \begin{subfigure}{0.4\textwidth}
    \centering
    \frame{\includegraphics[width=\textwidth]{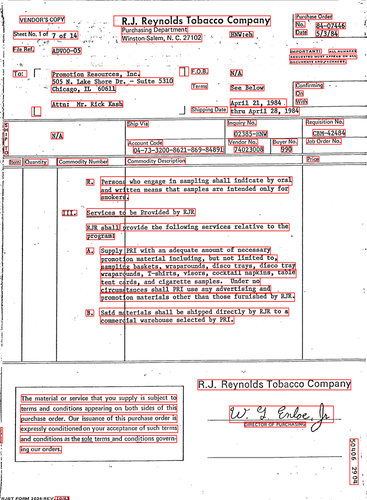}}
    \caption{Rendered output}
    \label{fig:cdip_sub2}
  \end{subfigure}

  \caption{A training sample for the layout-based task from IIT-CDIP}
  \label{fig:ocr_cdip}
\end{figure*}

\begin{figure*}[htbp]
  \centering

  \begin{subfigure}{0.4\textwidth}
    \centering
    \frame{\includegraphics[width=\textwidth]{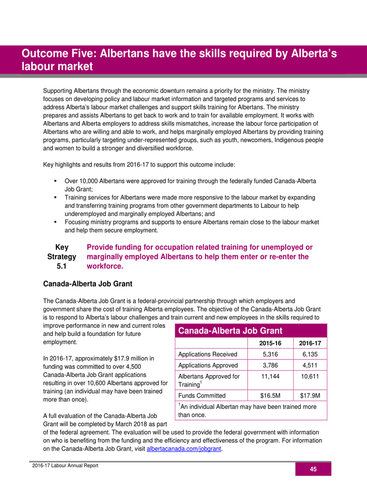}}
    \caption{Input}
    \label{fig:pdf_sub1}
  \end{subfigure}%
  ~
  \begin{subfigure}{0.4\textwidth}
    \centering
    \frame{\includegraphics[width=\textwidth]{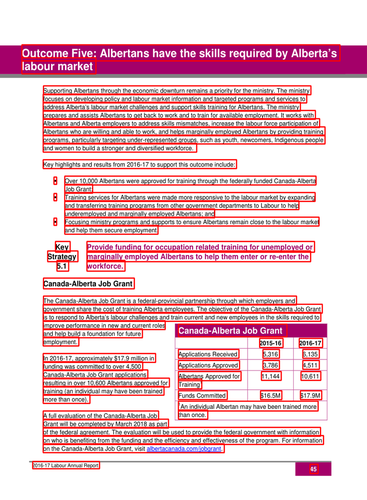}}
    \caption{Rendered output}
    \label{fig:pdf_sub2}
  \end{subfigure}

  \caption{A training sample for the layout-based task from PDFs}
  \label{fig:ocr_pdf}
\end{figure*}

\begin{figure*}[htbp]
  \centering

  \begin{subfigure}{0.4\textwidth}
    \centering
    \frame{\includegraphics[width=\textwidth]{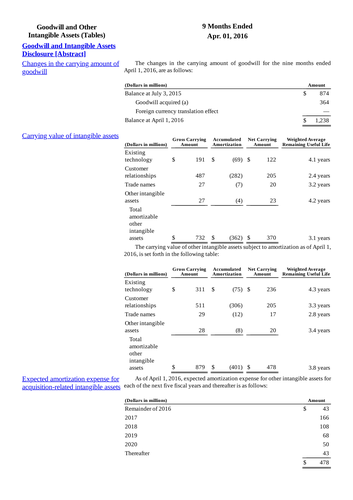}}
    \caption{Input}
    \label{fig:ocr_sec_input}
  \end{subfigure}%
  ~
  \begin{subfigure}{0.4\textwidth}
    \centering
    \frame{\includegraphics[width=\textwidth]{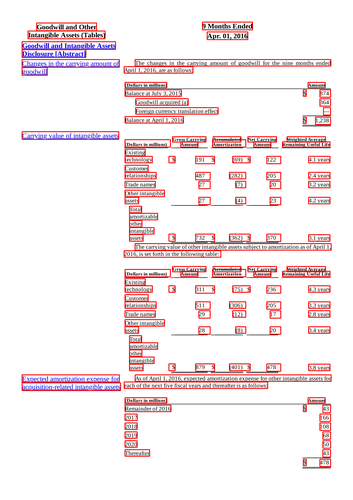}}
    \caption{Rendered output}
    \label{fig:ocr_sec_output}
  \end{subfigure}

  \caption{A training sample for the layout-based task from SECs}
  \label{fig:ocr_sec}
\end{figure*}

\begin{figure*}[htbp]
  \centering

  \begin{subfigure}{0.4\textwidth}
    \centering
    \frame{\includegraphics[width=\textwidth]{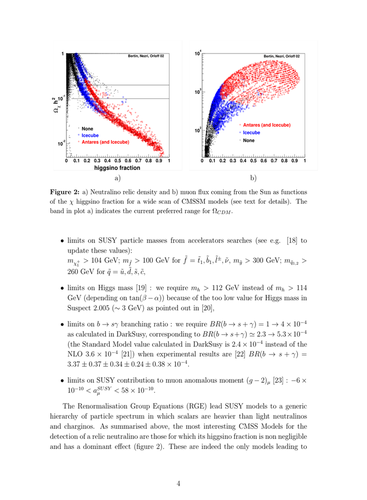}}
    \caption{Input}
    \label{fig:arxiv1_sub1}
  \end{subfigure}%
  ~
  \begin{subfigure}{0.4\textwidth}
    \centering
    \frame{\includegraphics[width=\textwidth]{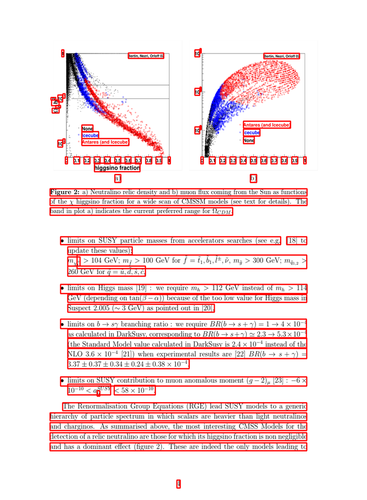}}
    \caption{Rendered output}
    \label{fig:arxiv1_sub2}
  \end{subfigure}

  \caption{A training sample for the layout-based task from arXiv papers (single-column)}
  \label{fig:ocr_arxiv1}
\end{figure*}

\begin{figure*}[htbp]
  \centering

  \begin{subfigure}{0.4\textwidth}
    \centering
    \frame{\includegraphics[width=\textwidth]{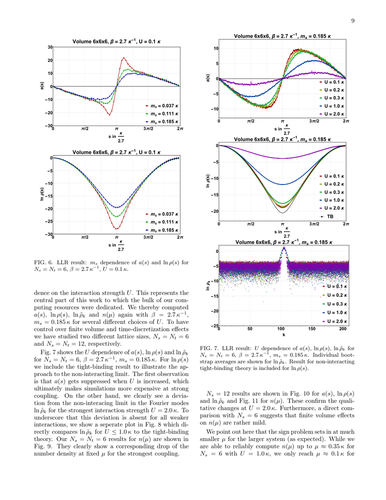}}
    \caption{Input}
    \label{fig:arxiv_sub1}
  \end{subfigure}%
  ~
  \begin{subfigure}{0.4\textwidth}
    \centering
    \frame{\includegraphics[width=\textwidth]{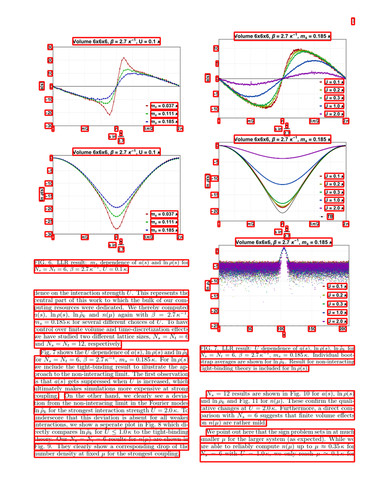}}
    \caption{Rendered output}
    \label{fig:arxiv_sub2}
  \end{subfigure}

  \caption{A training sample for the layout-based task from arXiv papers (two-column)}
  \label{fig:ocr_arxiv2}
\end{figure*}

\begin{figure*}[htbp]
  \centering

  \begin{subfigure}{0.4\textwidth}
    \centering
    \frame{\includegraphics[width=\textwidth]{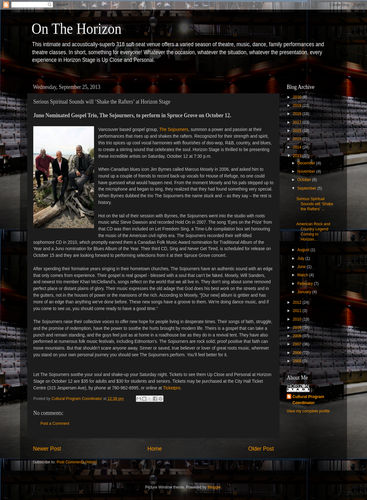}}
    \caption{Input}
    \label{fig:screen_sub1}
  \end{subfigure}%
  ~
  \begin{subfigure}{0.4\textwidth}
    \centering
    \frame{\includegraphics[width=\textwidth]{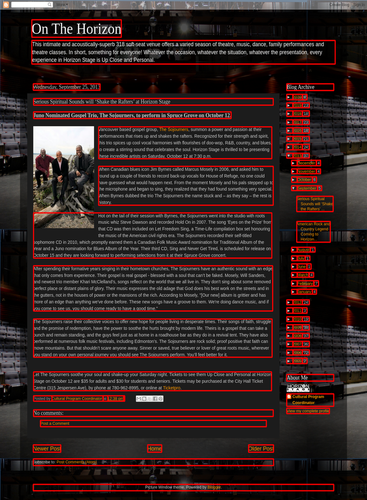}}
    \caption{Rendered output}
    \label{fig:screen_sub2}
  \end{subfigure}

  \caption{A training sample for the layout-based task from web screenshots}
  \label{fig:ocr_screen}
\end{figure*}

\begin{figure*}[htbp]
  \centering

  \begin{subfigure}{0.4\textwidth}
    \centering
    \frame{\includegraphics[width=\textwidth]{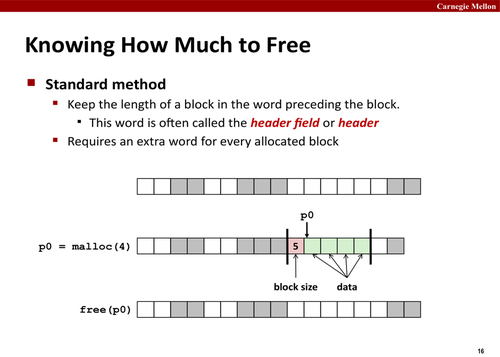}}
    \caption{Input}
    \label{fig:ppt_sub1}
  \end{subfigure}%
  ~
  \begin{subfigure}{0.4\textwidth}
    \centering
    \frame{\includegraphics[width=\textwidth]{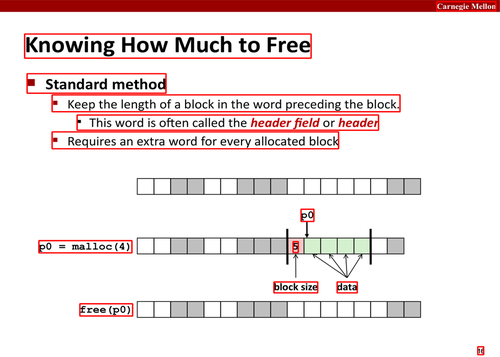}}
    \caption{Rendered output}
    \label{fig:ppt_sub2}
  \end{subfigure}

  \caption{A training sample for the layout-based task from PowerPoint slides}
  \label{fig:ocr_ppt}
\end{figure*}

\begin{figure*}[htbp]
  \centering

  \begin{subfigure}{0.4\textwidth}
    \centering
    \frame{\includegraphics[width=\textwidth]{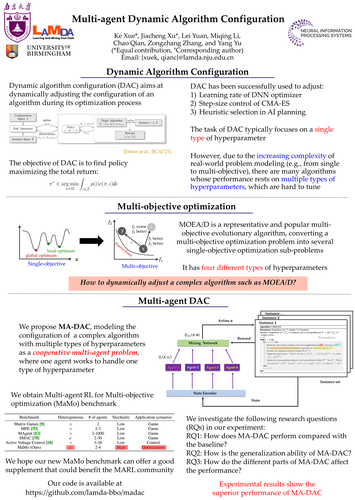}}
    \caption{Input}
    \label{fig:poster_sub1}
  \end{subfigure}%
  ~
  \begin{subfigure}{0.4\textwidth}
    \centering
    \frame{\includegraphics[width=\textwidth]{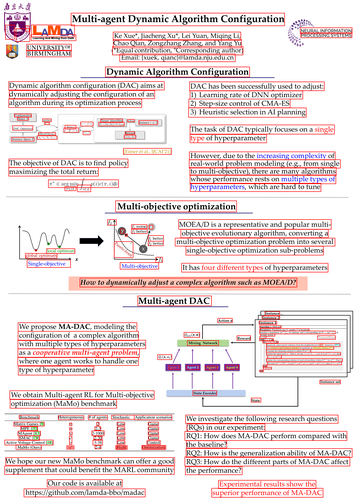}}
    \caption{Rendered output}
    \label{fig:poster_sub2}
  \end{subfigure}

  \caption{A training sample for the layout-based task from posters}
  \label{fig:ocr_poster}
\end{figure*}

\begin{figure*}[htbp]
  \centering

  \begin{subfigure}{0.4\textwidth}
    \centering
    \frame{\includegraphics[width=\textwidth]{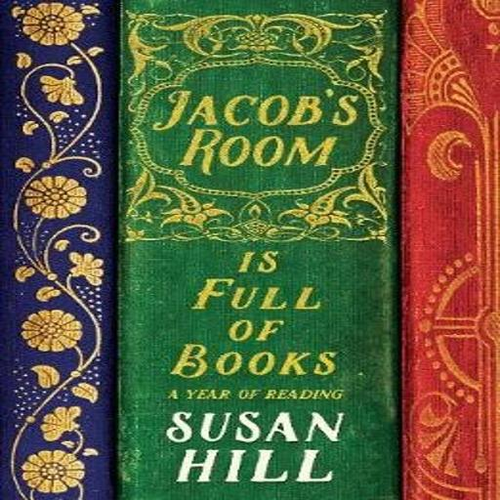}}
    \caption{Input}
    \label{fig:mario_sub1}
  \end{subfigure}%
  ~
  \begin{subfigure}{0.4\textwidth}
    \centering
    \frame{\includegraphics[width=\textwidth]{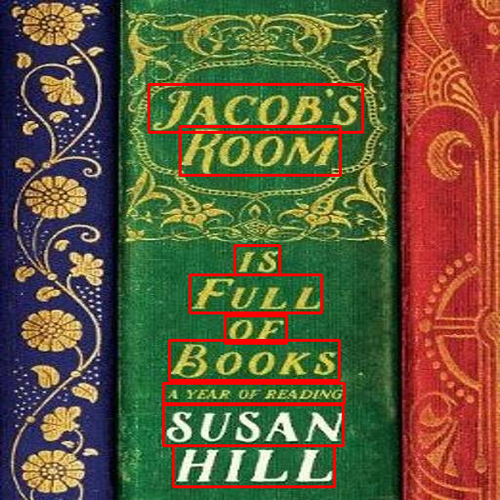}}
    \caption{Rendered output}
    \label{fig:mario_sub2}
  \end{subfigure}

  \caption{A training sample for the layout-based task from Mario-10M}
  \label{fig:ocr_mario}
\end{figure*}

\begin{figure*}[htbp]
  \centering

  \begin{subfigure}{0.4\textwidth}
    \centering
    \frame{\includegraphics[width=\textwidth]{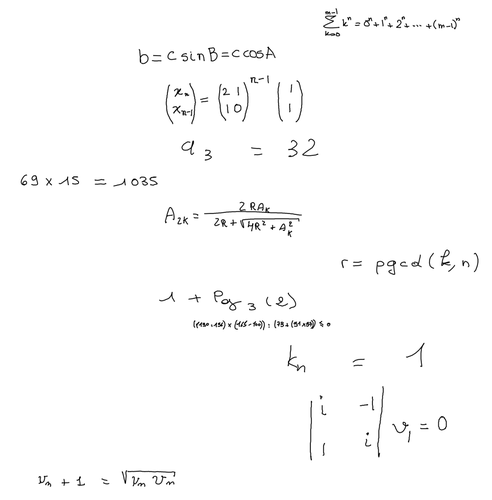}}
    \caption{Input}
    \label{fig:math_sub1}
  \end{subfigure}%
  ~
  \begin{subfigure}{0.4\textwidth}
    \centering
    \frame{\includegraphics[width=\textwidth]{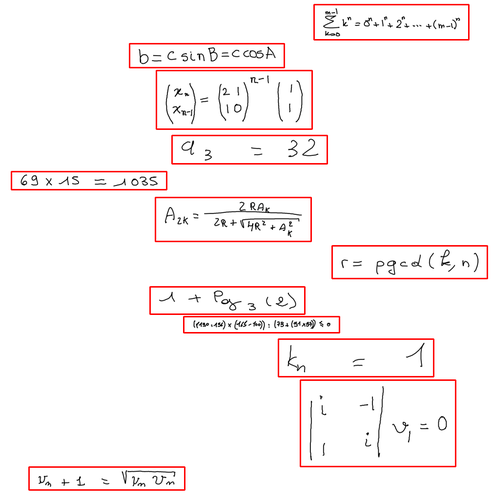}}
    \caption{Rendered output}
    \label{fig:math_sub2}
  \end{subfigure}

  \caption{A training sample for the layout-based task from mathematical}
  \label{fig:ocr_math}
\end{figure*}

\begin{figure*}[htbp]
  \centering

  \begin{subfigure}{0.4\textwidth}
    \centering
    \frame{\includegraphics[width=\textwidth]{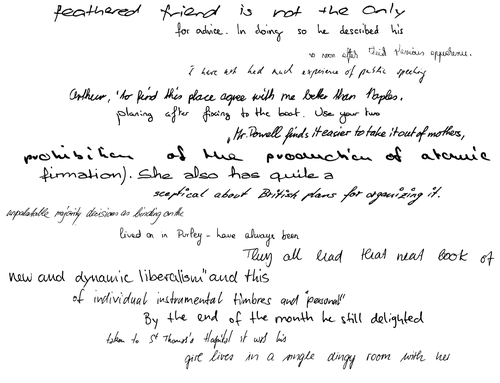}}
    \caption{Input}
    \label{fig:hw_sub1}
  \end{subfigure}%
  ~
  \begin{subfigure}{0.4\textwidth}
    \centering
    \frame{\includegraphics[width=\textwidth]{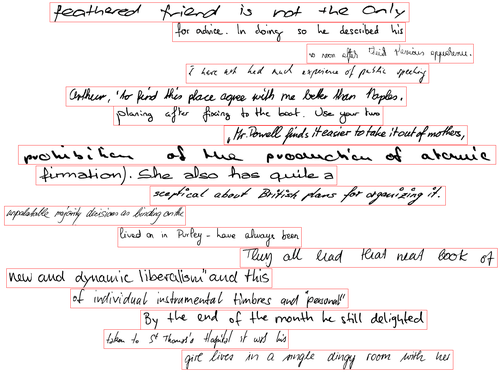}}
    \caption{Rendered output}
    \label{fig:hw_sub2}
  \end{subfigure}

  \caption{A training sample for the layout-based task from handwrittens}
  \label{fig:ocr_handwritten}
\end{figure*}

\begin{figure*}[htbp]
  \centering
  \begin{subfigure}{0.4\textwidth}
    \centering
    \frame{\includegraphics[width=\textwidth]{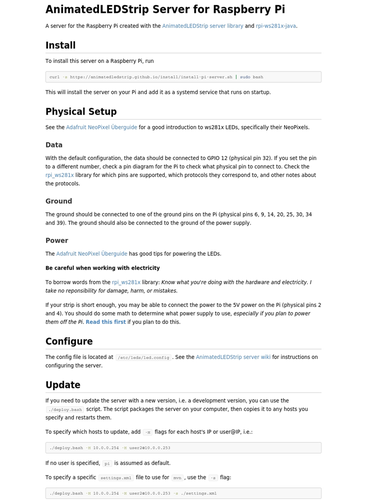}}
    \caption{Input}
    \label{fig:readme_sub1}
  \end{subfigure}%
  ~
  \begin{subfigure}{0.4\textwidth}
    \centering
    \frame{\includegraphics[width=\textwidth]{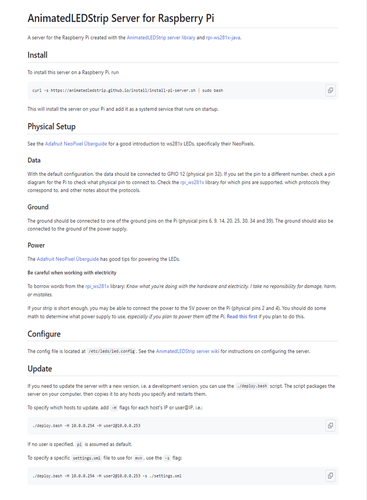}}
    \caption{Rendered output}
    \label{fig:readme_sub2}
  \end{subfigure}

  \caption{A training sample for the markup-based task from README}
  \label{fig:md_readme}
\end{figure*}

\begin{figure*}[htbp]
  \centering
  \begin{subfigure}{0.4\textwidth}
    \centering
    \frame{\includegraphics[width=\textwidth]{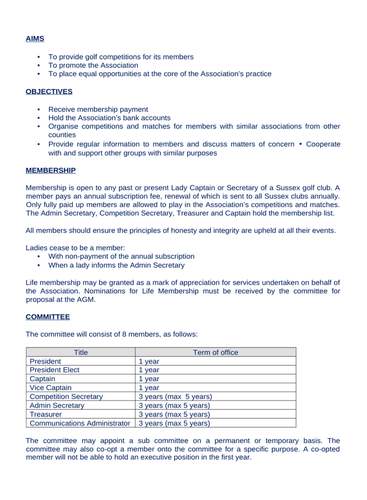}}
    \caption{Input}
    \label{fig:docx_sub1}
  \end{subfigure}%
  ~
  \begin{subfigure}{0.4\textwidth}
    \centering
    \frame{\includegraphics[width=\textwidth]{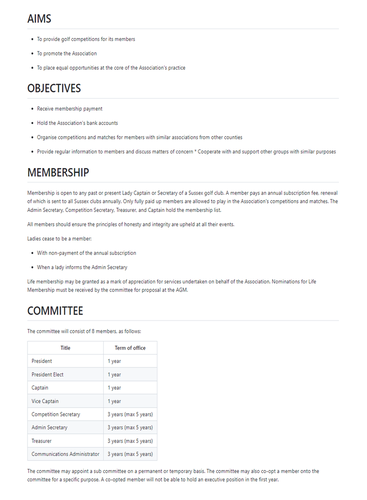}}
    \caption{Rendered output}
    \label{fig:docx_sub2}
  \end{subfigure}

  \caption{A training sample for the markup-based task from DOCX}
  \label{fig:md_docx}
\end{figure*}

\begin{figure*}[htbp]
  \centering
  \begin{subfigure}{0.4\textwidth}
    \centering
    \frame{\includegraphics[width=\textwidth]{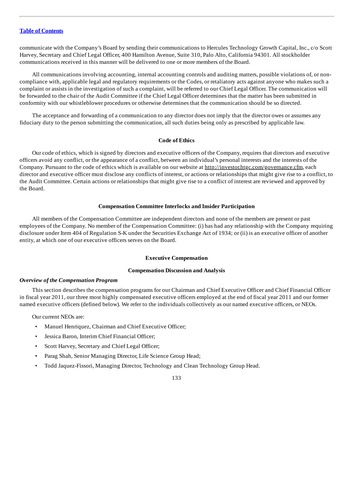}}
    \caption{Input}
    \label{fig:md_sec_input}
  \end{subfigure}%
  ~
  \begin{subfigure}{0.4\textwidth}
    \centering
    \frame{\includegraphics[width=\textwidth]{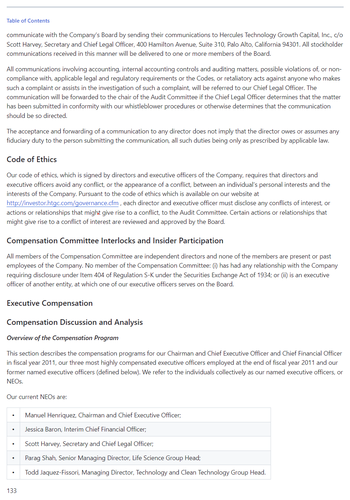}}
    \caption{Rendered output}
    \label{fig:md_sec_output}
  \end{subfigure}

  \caption{A training sample for the markup-based task from SEC}
  \label{fig:md_sec}
\end{figure*}

\begin{figure*}[htbp]
  \centering
  \begin{subfigure}{0.4\textwidth}
    \centering
    \frame{\includegraphics[width=\textwidth]{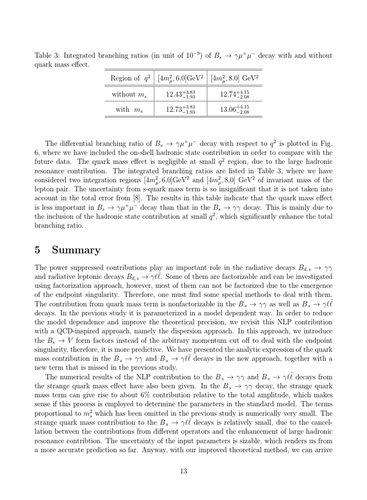}}
    \caption{Input}
    \label{fig:latex1_sub1}
  \end{subfigure}%
  ~
  \begin{subfigure}{0.4\textwidth}
    \centering
    \frame{\includegraphics[width=\textwidth]{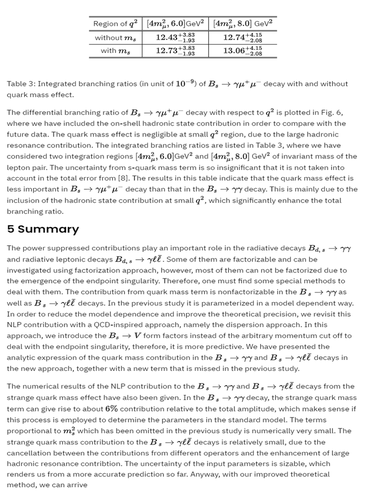}}
    \caption{Rendered output}
    \label{fig:latex1_sub2}
  \end{subfigure}

  \caption{A training sample for the markup-based task from \LaTeX\ (single-column)}
  \label{fig:md_latex1}
\end{figure*}

\begin{figure*}[htbp]
  \centering
  \begin{subfigure}{0.4\textwidth}
    \centering
    \frame{\includegraphics[width=\textwidth]{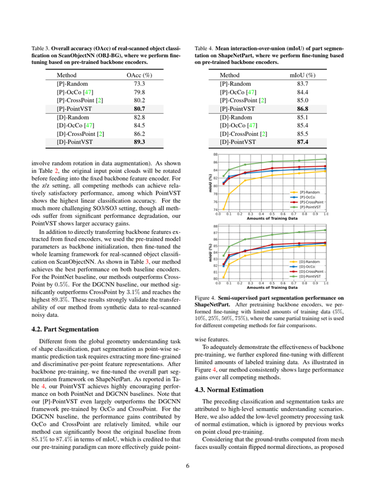}}
    \caption{Input}
    \label{fig:latex_sub1}
  \end{subfigure}%
  ~
  \begin{subfigure}{0.4\textwidth}
    \centering
    \frame{\includegraphics[width=\textwidth]{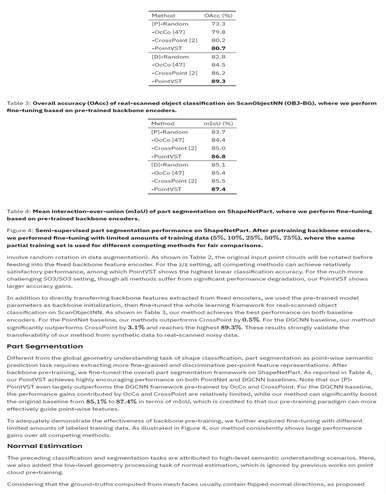}}
    \caption{Rendered output}
    \label{fig:latex_sub2}
  \end{subfigure}

  \caption{A training sample for the markup-based task from \LaTeX\ (two-column)}
  \label{fig:md_latex2}
\end{figure*}

\begin{figure*}[htbp]
  \centering
  \begin{subfigure}{0.4\textwidth}
    \centering
    \frame{\includegraphics[width=\textwidth]{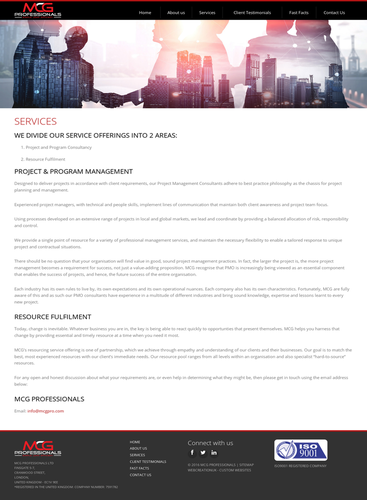}}
    \caption{Input}
    \label{fig:html_sub1}
  \end{subfigure}%
  ~
  \begin{subfigure}{0.4\textwidth}
    \centering
    \frame{\includegraphics[width=\textwidth]{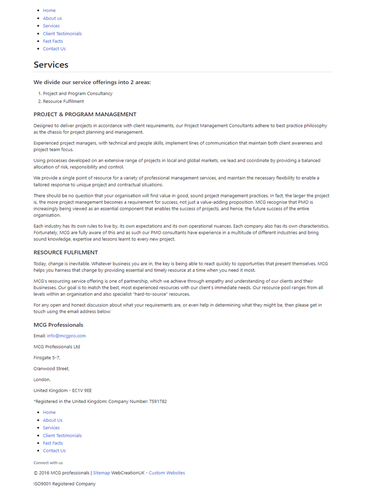}}
    \caption{Rendered output}
    \label{fig:html_sub2}
  \end{subfigure}

  \caption{A training sample for the markup-based task from HTMLs}
  \label{fig:md_screen}
\end{figure*}

\newpage

\end{document}